\def\paperlanguage{}
\pdfoutput=1 % for arxiv

\documentclass[letterpaper, 10 pt, conference]{ieeeconf}  % Comment this line out if you need a4paper

\usepackage{bm}
\usepackage{cite}
\usepackage{balance}
\include{preamble}
\usepackage{siunitx}

\IEEEoverridecommandlockouts                              % This command is only needed if
\title{\LARGE \textbf
  {
    \switchlanguage%
    {%
      Fundamental Three-Dimensional Configuration\\of Wire-Wound Muscle-Tendon Complex Drive
    }%
    {%
      ワイヤ巻取式筋腱複合体駆動の\\基礎的な三次元的構成の開発
    }%
  }
}

\author{Yoshimoto Ribayashi$^{1}$, Yuta Sahara$^{1}$, Shogo Sawaguchi$^{1}$, Kazuhiro Miyama$^{1}$,\\Akihiro Miki$^{1}$, Kento Kawaharazuka$^{1}$, Kei Okada$^{1}$, and Masayuki Inaba$^{1}$% <-this % stops a space
  \thanks{$^{1}$ The authors are with the Department of Mechano-Informatics, Graduate School of Information Science and Technology, The University of Tokyo, 7-3-1 Hongo, Bunkyo-ku, Tokyo, 113-8656, Japan.
    {\texttt\small [ribayashi, sahara, sawaguchi, miyama, miki, kawaharazuka, k-okada, inaba]@jsk.t.u-tokyo.ac.jp}
  }
}
\begin{document}

\maketitle
\thispagestyle{empty}
\pagestyle{empty}

%%%%%%%%%%%%%%%%%%%%%%%%%%%%%%%%%%%%%%%%%%%%%%%%%%%%%%%%%%%%%%%%%%%%%%%%%%%%%%%%
\begin{abstract}
  \switchlanguage%
  {%
    For robots to become more versatile and expand their areas of application, their bodies need to be suitable for contact with the environment.
    When the human body comes into contact with the environment, it is possible for it to continue to move even if the positional relationship between muscles or the shape of the muscles changes.
    We have already focused on the effect of geometric deformation of muscles and proposed a drive system called wire-wound Muscle-Tendon Complex (ww-MTC), an extension of the wire drive system.
    %Using a robot with a two-dimensional configuration, we confirmed the following effects: suppression of wire loosening, interference, and wear, robustness in environmental contact, and muscular appearance.
    %On the other hand, there were some problems, such as too large muscle expansion, which caused the muscles to inhibit each other's movement, and movement was limited to the plane.
    Our previous study using a robot with a two-dimensional configuration demonstrated several advantages: reduced wire loosening, interference, and wear; improved robustness during environmental contact; and a muscular appearance.
    However, this design had some problems, such as excessive muscle expansion that hindered inter-muscle movement, and confinement to planar motion.
    %In this study, we develop the ww-MTC into a three-dimensional geometry and show a fundamental construction method of a muscle exterior that expands gently and can be contacted over the entire surface.
    In this study, we develop the ww-MTC into a three-dimensional shape.
    We present a fundamental construction method for a muscle exterior that expands gently and can be contacted over its entire surface.
    We also apply the three-dimensional ww-MTC to a 2-axis 3-muscle robot, and confirm that the robot can continue to move while adapting to its environment.
  }%
  {%
    ロボットがより汎用的となり活躍の場を広げるためには, 環境接触に適した身体の構成が必要不可欠である.
    人間は身体と環境が接触する際, 筋肉同士の位置関係や形状が変化しても動作を継続することができている.
    我々は既に筋肉の幾何学的変形の効果に注目し, ワイヤ駆動を発展させたwire-wound Muscle-Tendon Complex (ww-MTC)という駆動方式を提案した\cite{MTC-2D:Ribayashi:HUMANOIDS2023}.
    2次元的構成のロボットを用いて, ワイヤの緩み, 干渉, 摩耗の抑制, 環境接触時のロバスト性, 筋肉質な見た目といった効果を確認した.
    一方で筋の膨張が大きすぎるために筋同士が互いに動きを阻害している, 動作が平面内に限られるといった課題があった.
    本研究ではww-MTCを3次元的形状に発展させ, ゆるやかに膨張し表面全体が覆われた筋外装の基本的な構成法を示す.
    また, 2軸3筋のロボットに3次元的なww-MTCを適用し, 環境に馴染みながら動作を継続できることを確認した.
  }%
\end{abstract}

\section{Introduction} \label{sec:introduction}
\switchlanguage%
{%
  %Humanoid robots have been actively developed by companies in recent years\cite{GR-1, talos, apollo, neo}.
  Humanoid robots have been actively developed by companies in recent years\cite{GR-1, talos, apollo, neo}, achieving dexterous manipulation and stable locomotion.
  Most of the robots presented have rigid bodies, and they deal with environmental contact only at the end-effectors, such as hands and feet.
  As for contact with other parts of the body, although there are some examples of studies on multi-point contact\cite{hiraoka2019whole, ruscelli2020multi, ferrari2023multi} and disturbances applied during push recovery motions\cite{pratt2006capture, stephens2010push, semwal2017robust}, the topics covered are limited.
  For example, complex movements involving whole-body surface contact, such as holding with arms and torso, or pushing aside obstacles with the body while both hands are occupied, are challenging to achieve. 
  %For example, complex and challenging movements involving whole-body surface contact, such as holding with arms and torso, or pushing aside obstacles with the body while both hands are occupied, are challenging to achieve.
  If the robot tries to force its body to move when it is in contact with the environment, its rigid body may damage the surrounding environment.
  In order for robots to play a more active role in society, it is necessary for robots to handle a broader range of environmental contacts and to have a body suitable for such contacts.

  There are soft humanoids that fit into the environment, such as those using a wire drive\cite{wittmeier2013toward, jantsch2013anthrob, iros2019:kawaharazuka:musashi} and those using pneumatic artificial muscles\cite{Humanoids2012:mizuuchi:pneumatic, AR2018:hitzmann:anthropomorphic}.
  Although each drive method has its own advantages and disadvantages, we focus on the wire drive in this research.
  The major advantage of the wire drive is that it exerts a large force and its high controllability due to the winding by the motor.
  However, handling environmental contact with the entire body is challenging because of unexpected interference and loosening of the wires in a body configuration with exposed wires.
  Humanoids need bodies that can adapt to the environment like humans do.
  They must also be able to operate while making good use of the environment by contacting all body surfaces and continuing to operate safely in the face of environmental disturbances.

\begin{figure}[t]
 \centering
 \includegraphics[width=1.00\columnwidth]{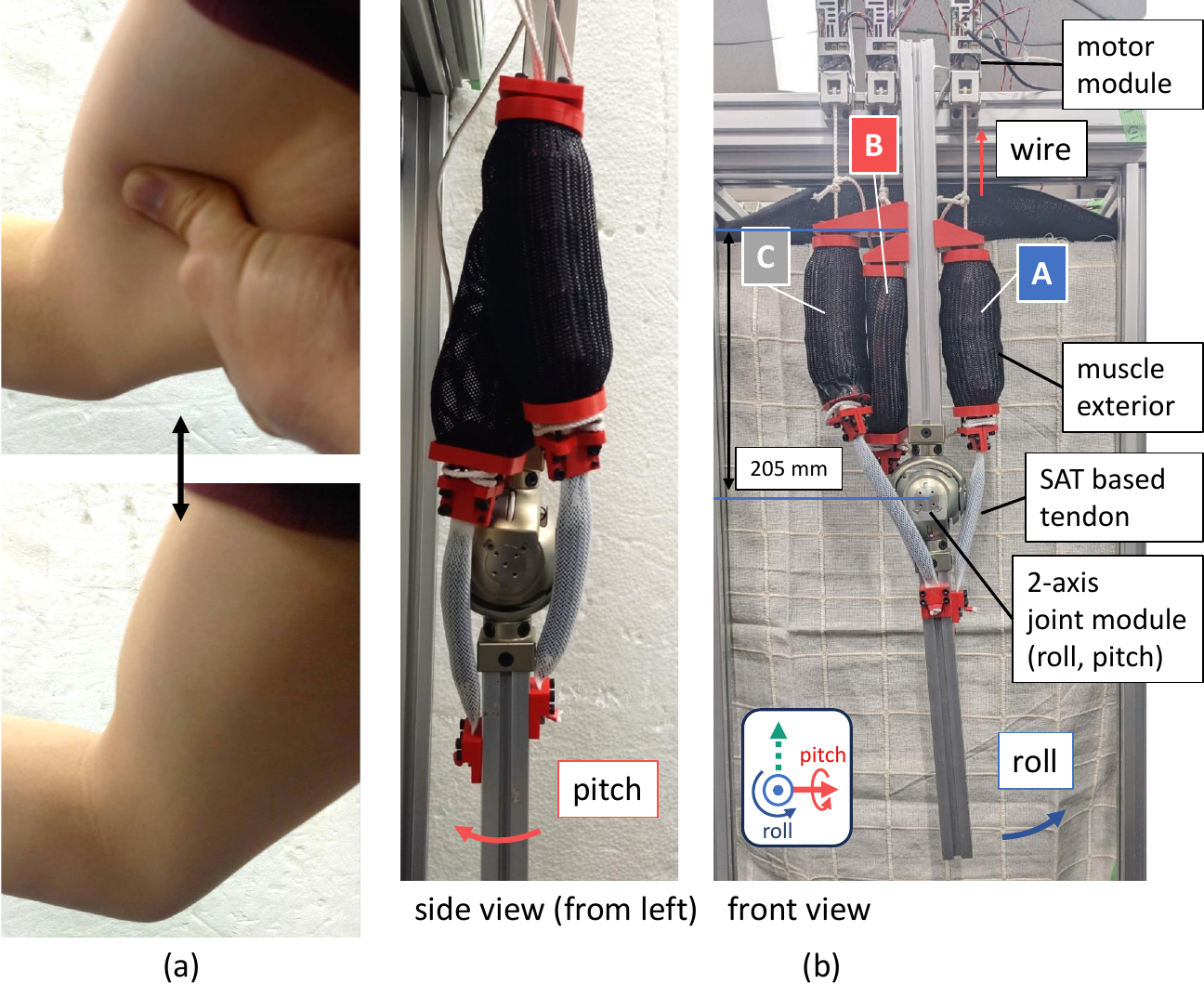}
 \vspace*{-4mm}
  \caption{(a) Muscle-tendon complexes adjust their shape and internal spatial relationships when encountering contact, enabling continued movement while adapting to the environment. They return to their original position after contact with the environment ends. (b) 2-axis 3-muscle robot with three-dimensional configuration.}
  \label{fig:abst}
 \vspace*{-6mm}
\end{figure}

  %Humans' adaptability to such environmental contact is mainly due to the muscle-tendon complexes that drive the body.
  %Muscle-tendon complexes contribute to the body's flexibility.
  %They can even change the positional relationship of their muscles and tendons in response to the external shape and external forces they come into contact with and adapt to them while continuing to perform their movements.
  %Moreover, they quickly return to their original positions and shapes when the contact is over (\figref{fig:abst}(a)).
  %In this study, we focus on this point and aim to develop a body driving system with high adaptability in contact with the environment.

  %claude
  %Humans' adaptability to environmental contact is primarily due to the muscle-tendon complexes that drive the body.
  %Muscle-tendon complexes significantly contribute to the body's flexibility.
  %In response to external contact, they can change the positional relationship of muscles and tendons, adapting to the environment while continuing to perform movements.
  The human body's remarkable adaptability to environmental contact is primarily due to its muscle-tendon complexes.
  These complexes not only provide flexibility but also enable dynamic adaptation to external forces.
  When encountering environmental contact, muscle-tendon complexes can adjust their shape and spatial relationships between muscles and tendons.
  This adaptability allows for continued movement while conforming to new environmental conditions.
  Moreover, they quickly return to their original positions and shapes when contact ceases (\figref{fig:abst}(a)).
  In this study, we focus on these adaptive characteristics to develop a body-driving system that is highly adaptable to environmental contact.

\begin{figure}[t]
  \centering
      \includegraphics[width=1.0\columnwidth]{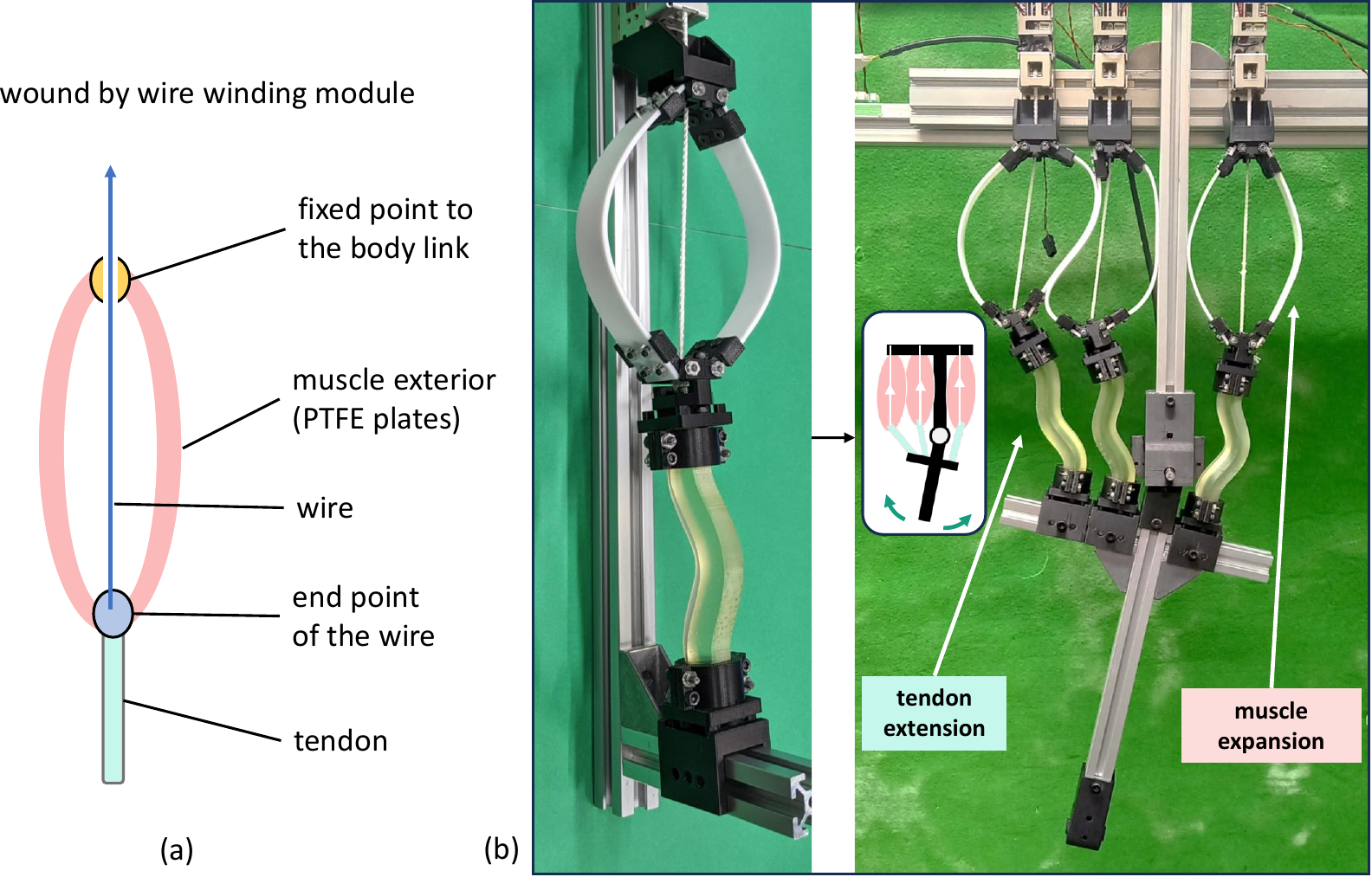}
 \vspace*{-4mm}
  \caption{ (a) Conceptual diagram of wire-wound Muscle-Tendon Complex drive\cite{MTC-2D:Ribayashi:HUMANOIDS2023}. (b) 1-axis 3-muscle robot with two-dimensional configuration\cite{MTC-2D:Ribayashi:HUMANOIDS2023}.}
  \label{fig:2D}
 \vspace*{-4mm}
\end{figure}

  We have already proposed a drive system called the wire-wound Muscle-Tendon Complex (ww-MTC), which incorporates muscle-tendon deformation into the wire drive\cite{MTC-2D:Ribayashi:HUMANOIDS2023}(\figref{fig:2D}).
  The concept of incorporating geometrical shape changes of muscles in a two-dimensional configuration is realized, and advantages such as prevention of wire loosening, securing of moment arms, internal protection against environmental contact, and muscular appearance are confirmed.
  This method is similar to pneumatic artificial muscles such as the McKibben type\cite{TRO1996:Ching:McKibben, Act2022:Kalita:McKibben} in that it expands with contraction. However, since the tension is provided by an internal wire, the shape can be changed in various ways according to the design of the muscle exterior.
  This study concludes that the incorporation of muscle expansion is promising for body configurations suitable for environmental contact.
  On the other hand, this configuration has the problem of too large expansions of the muscle parts inhibiting each other's movement.
  In addition, developing a three-dimensional muscle exterior that has no gaps on its surface and can expand and deform has been a major challenge.

  In order to use muscle-tendon complex drives in three-dimensional movements and body configurations, we develop muscle exterior and tendon elements with more compact shapes and entirely covered surfaces.
  In particular, for muscles that expand and deform, we propose a structure consisting of a frame made of arch-shaped material that defines the deformation and a braided sleeve that covers the surface.
  In addition, the proposed structure's effectiveness is verified by performing motion experiments with environmental contact on a 2-axis, 3-muscle robot using a three-dimensional ww-MTC.

  The main contributions of this study are as follows:
  \begin{itemize}
  %\item This study shed light on the high adaptability of deformable muscles in contact with the environment.
  %\item We proposed a structure that deforms three-dimensionally under wire tension, and confirmed the relationship between contraction distance and expansion through theoretical calculations.
  %\item We have verified that the structure works without any problem, even if the position and shape of the muscles change during contact with the environment, using 3D ww-MTC on an actual robot.
  \item This study shed light on the inherent adaptability of deformable biological muscles during environmental contact and aimed to achieve it in robotic systems.
  \item We proposed a structure that deforms three-dimensionally under wire tension, and confirmed the relationship between contraction distance and expansion through theoretical calculations.
  %\item We have verified that the system maintains operation even if the position and shape of the muscles change during environmental contact, using 3D ww-MTC on an actual robot.
  \item We have verified that the robot using 3D ww-MTC maintains its operation even if the position and shape of the muscles change during environmental contact.
  \end{itemize}
}%
{%
  近年ヒューマノイドロボットは企業でも盛んに開発されている\cite{GR-1, talos, apollo, neo}. %tesla
  既に発表されてきた様々なロボットは剛な身体を持ち, 手先や足先といったエンドエフェクタでのみ環境接触を扱っているものが多い.
  他の部位での接触に関しては, 動作計画時の多点接触\cite{hiraoka2019whole, ruscelli2020multi, ferrari2023multi}やプッシュリカバリー動作\cite{pratt2006capture, stephens2010push, semwal2017robust}で加えられる外乱などの研究例はあるものの, 扱われている内容は限定的である.
  例えば腕と胴体を使った抱え込みや, 両手がふさがった状態で身体で障害物を押しのけるといった, 全身面の接触を伴う動作は複雑で難しい.
  環境接触時に無理に動作を行おうとすれば, 剛な身体では周辺の環境を傷つけてしまうこともある.
  ロボットがより社会で活躍するためには, より幅広い種類の環境接触を扱うこと, それに適した身体を有することが必要である.

  環境に馴染む柔らかいヒューマノイドとしてはワイヤ駆動を用いたもの\cite{wittmeier2013toward, jantsch2013anthrob, iros2019:kawaharazuka:musashi}や空気圧人工筋肉を用いたもの\cite{Humanoids2012:mizuuchi:pneumatic, AR2018:hitzmann:anthropomorphic}などがある.
  駆動方式ごとに利点と欠点があるが, 本研究ではワイヤ駆動を対象とする.
  ワイヤ駆動はモータの巻取りにより発揮力が大きく, 制御性も高いことが大きな利点である.
  しかし, ワイヤが剥き出しになった身体構成では, 予期せぬワイヤの干渉や緩みが生じるため, 身体全体での環境接触を扱うことは難しい.
  %しかし
  ヒューマノイドは人間のように環境に馴染む身体を持った上で, 全身の表面を環境に接触させて上手く利用しながら動作したり, 身体への環境接触がありながらも安全に動作を継続できたりするようになる必要がある.

  人間の場合, このような環境接触における適応性は身体を駆動する筋腱によるところが大きい.
  %生物を駆動する身体構成要素としての
  筋腱複合体は身体の柔軟性に寄与するとともに, 接触する外部形状や外力に対して筋腱の位置関係さえも変化させて馴染みながら, 動作を継続することができる.
  その上, 接触が終わると素早く元の位置関係と形状に戻る (\figref{fig:abst}(a)). %という特徴を持っている.
  本研究ではこの点に注目し, 環境接触における高い適応性を持つ身体の駆動システムを目指す.

  我々は既に, ワイヤ駆動に筋腱の変形を取り入れたwire-wound Muscle-Tendon Complex (ww-MTC)という駆動システムを提案した\cite{MTC-2D:Ribayashi:HUMANOIDS2023}(\figref{fig:2D}).
  二次元的な構成で筋肉の幾何学的形状変化を取り入れるというコンセプトを具現化し, ワイヤの緩み防止やモーメントアーム確保, 環境接触時の内部保護, 筋肉質な見た目といった利点を確認している.
  この手法は収縮に伴い膨張する点でMcKibben型などの空気圧人工筋肉\cite{TRO1996:Ching:McKibben, Act2022:Kalita:McKibben}と類似するが, 張力を内部のワイヤが担うため, 形状を筋外装の設計により様々に変化させられることが特徴である.
  この研究により筋の膨張を取り入れることは環境接触に適した身体構成のために有望であると結論づけられた.
  一方この構成では筋に相当する部分の大きすぎる膨張変形が互いの動きを阻害するという問題があった.
  また三次元的な形状の筋外装で, 表面に隙間がなくかつ膨張変形を実現するものの開発は大きな課題であった.

  そこで本研究では筋腱複合体駆動を, 三次元的な動きと身体構成においても使用するために, よりコンパクトな形状で表面全体が覆われた筋外装と腱要素を開発する.
  特に膨張変形する筋について, 変形を規定するアーチ状の材を用いた骨組みと表面を覆うbraided sleeveからなる構造を提案する.
  % 他の膨張変形ソフトロボット
  また, 3次元的なww-MTCを用いた2軸3筋のロボットにおいて環境接触を伴う動作実験を行い, 有効性を検証した.

  本研究の主な貢献は以下である.
  \begin{itemize}
  \item 変形する筋肉が持つ環境接触における高い適応性に光を当てた.
  \item ワイヤ張力で三次元的な変形をする構造を提案し, 理論的な計算によって収縮距離と膨張の関係を確認した.
  %\item 筋肉の環境接触における高い適応性をロボットに取り入れた
  \item 3次元的なww-MTCにより, 環境接触時に筋の位置や形状が変化しても問題なく動作することを実機で検証した.
  \end{itemize}
}%

\begin{figure}[tbh]
  \centering
  \includegraphics[width=0.9\columnwidth]{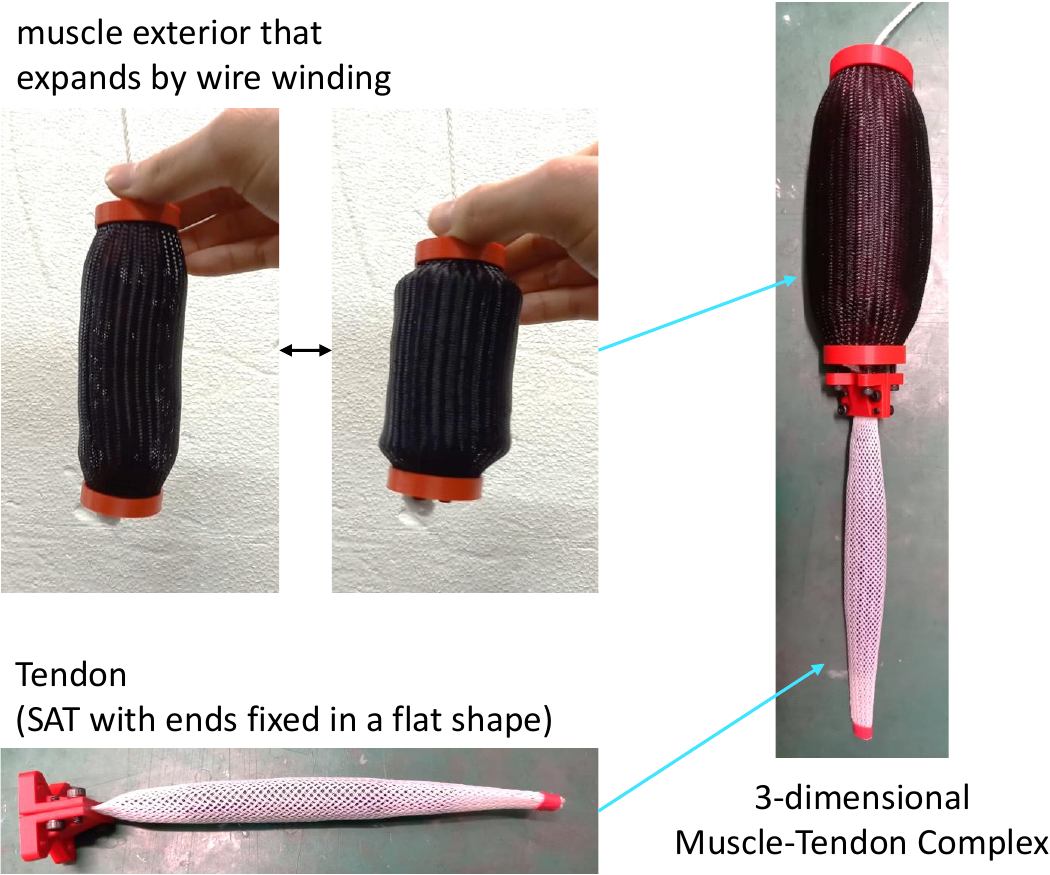}
  \caption{Three-dimensional ww-MTC consists of the muscle exterior and the tendon based on Stiffness Adjustable Tendon (SAT)\cite{robomech2003:shirai:sat}.}
  \label{fig:wip-mtc}
\end{figure}

\begin{figure}[tbh]
  \centering
  \includegraphics[width=0.9\columnwidth]{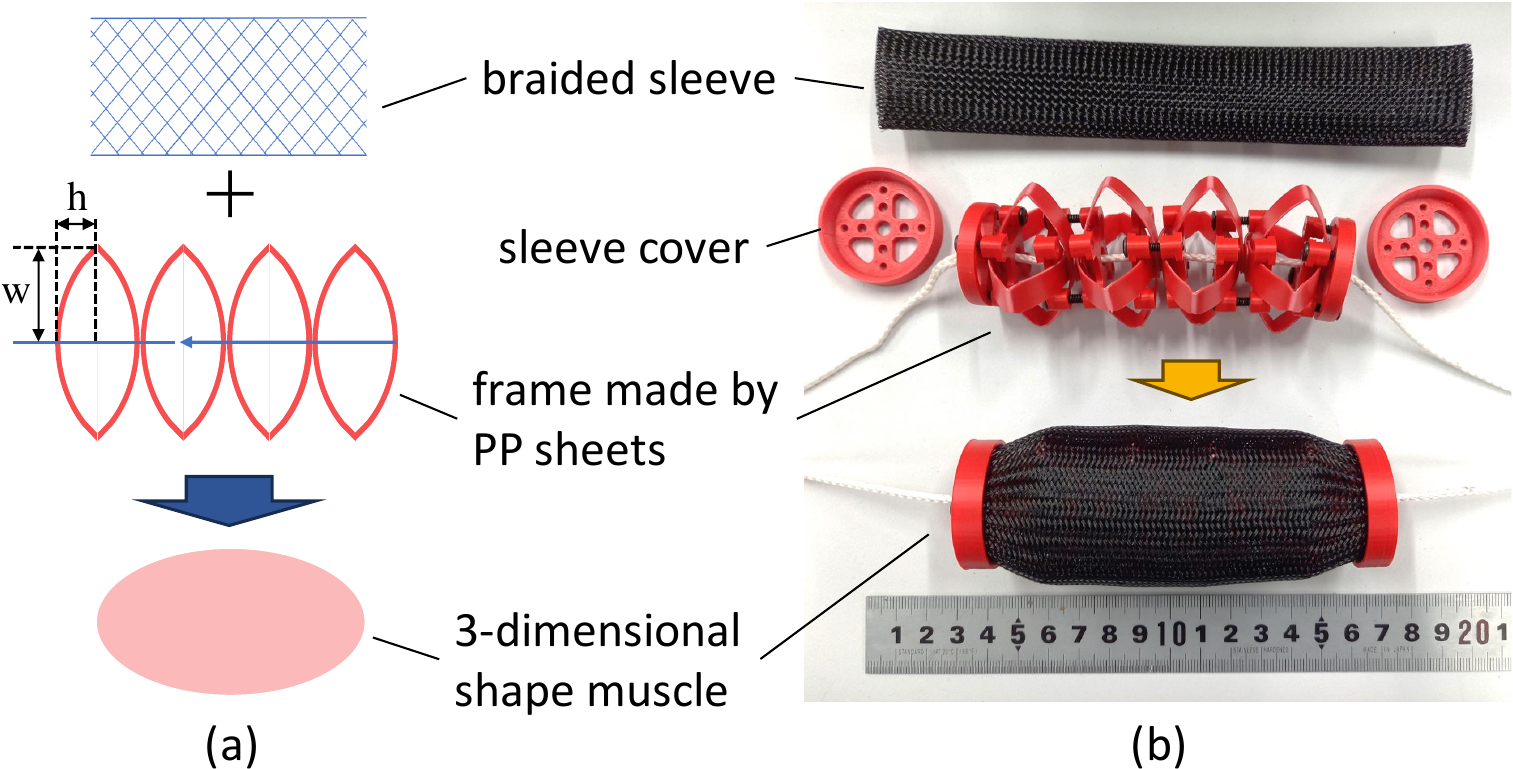}
  \caption{(a) Conceptual diagram of three-dimensional muscle. Muscle exterior is made by frame determining shape change and braided sleeve covering the surface. (b) Assembly of the muscle exterior.}
  \label{fig:exterior}
\end{figure}

\section{3D Configuration of a Wire-Wound Muscle-Tendon Complex Drive} \label{sec:mtc}

\switchlanguage%
{%
The three-dimensional muscle exterior requires the following.
\begin{itemize}
 \item The entire body surface is covered and contactable, i.e., usable in three-dimensional movements.
 \item  The muscle expansion is gentle and does not interfere with the movement of other muscles.
\end{itemize}

Tendons made of rubber are large compared to their strength, and have problems with deterioration and rupture\cite{MTC-2D:Ribayashi:HUMANOIDS2023}.
Therefore, the following tendon requirements are set.
\begin{itemize}
 \item The tendon is thin enough.
 \item The tendon is strong enough.
 \item The friction on the tendon's surface is small and does not interfere with contact at joints.
\end{itemize}

The level of detail in these tendon requirements depends on factors such as the size of the joint modules\cite{iros2019:kawaharazuka:musashi} and the performance of the motor modules\cite{iros2015:asano:module} used in this study.
Based on the above requirements, we propose a ww-MTC that can be used in a three-dimensional configuration (\figref{fig:wip-mtc}).

  For the winding of the wire, we use a wire winding module developed in our laboratory\cite{iros2015:asano:module}.
  This module can wind the wire by rotating the motor, and only the current is controlled in this study.
  The tension value is measured by the on-board load cell.
}%
{%
%二次元的な筋外装では膨張が大きいことで互いに接触する筋肉同士が過剰に押し合い, ロボットの動きに悪影響をもたらしていた.
三次元的な筋外装は以下を要件とする.
\begin{itemize}
 \item 表面全体が覆われており接触可能であること, すなわち三次元的な動きの中で使用することができること.
 \item 膨張がゆるやかであり, 他の筋の動きを阻害しないこと.
\end{itemize}

ゴムを用いた腱については, 強さに比較してサイズが大きく, また劣化して破断するといった問題が起きていた\cite{MTC-2D:Ribayashi:HUMANOIDS2023}.
そこで腱としては以下を要件とする.
\begin{itemize}
 \item 充分に細く構成できること.
 \item 充分に強いこと.
 \item 腱の表面の摩擦が小さく, 関節部での接触に支障をきたさないこと.
\end{itemize}

以上の要件に基づき, 三次元的な構成の中で用いることのできるww-MTCを提案する (\figref{fig:wip-mtc}).

  なお, ワイヤの巻取りには弊研究室で開発された筋モジュール\cite{iros2015:asano:module}を用いている.
  このモジュールはワイヤをモータの回転で巻き取ることが可能であり, 本研究では電流のみを制御する.
  張力値は搭載されているロードセルにより計測している.
}%

\begin{figure}[t]
  \centering
  \includegraphics[width=1.0\columnwidth]{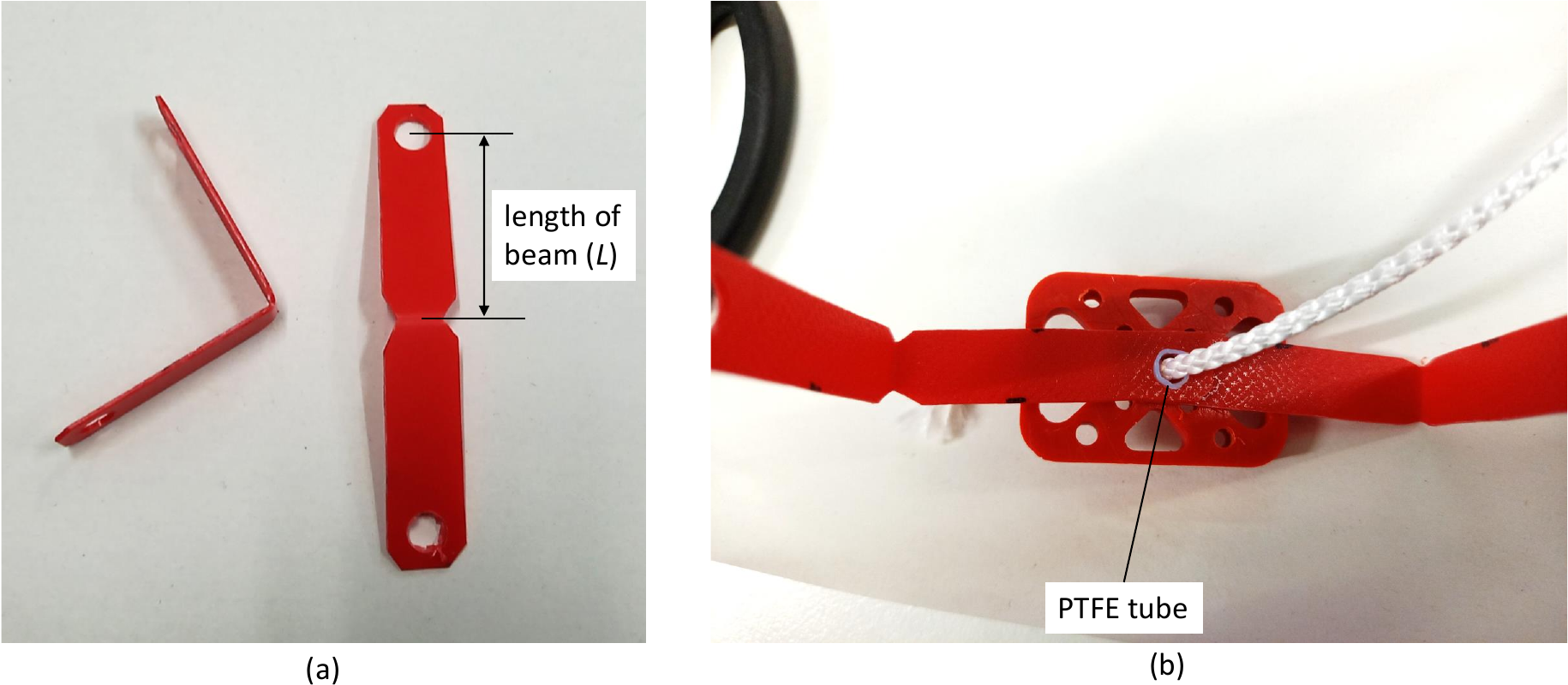}
  \caption{The PP sheet is processed as shown in (a). The distance from the hole through which the wire passes to the seam corresponds to the length of the beam and is an important parameter that defines the deformation of the muscle exterior. (b) PTFE tube is inserted to prevent wires from getting stuck.}
  \label{fig:PP}
\end{figure}

\begin{figure}[t]
 \centering
 \includegraphics[width=1.0\columnwidth]{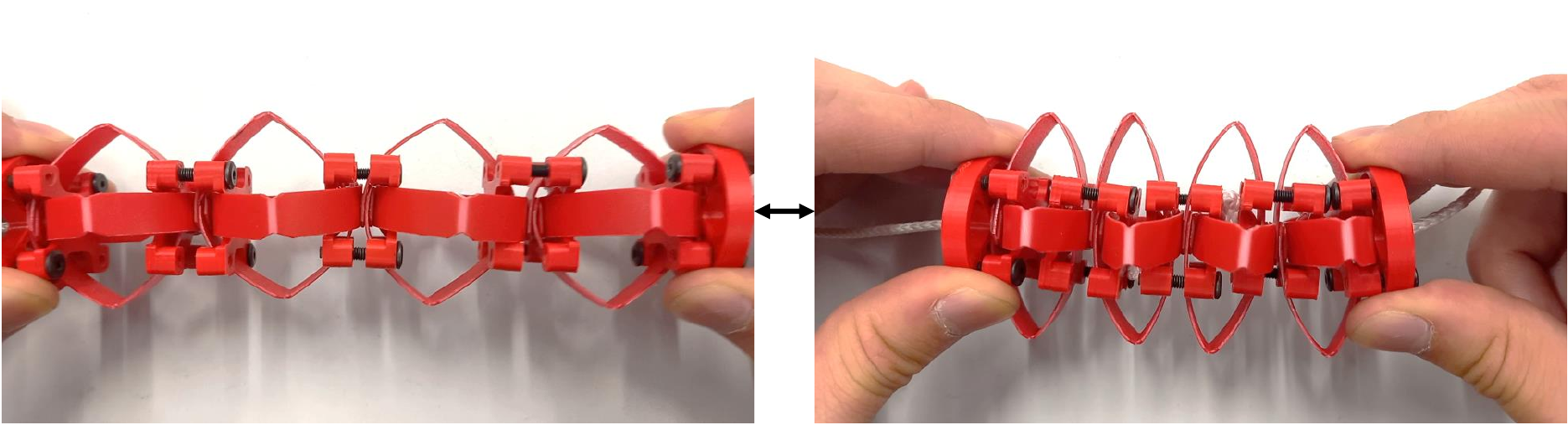}
 %\vspace*{-6mm}
  \caption{The image shows the deformation of the frame. The width becomes a little larger as the length becomes shorter.}
  \label{fig:length}
 %\vspace*{-4mm}
\end{figure}

\subsection{Muscle Exterior} \label{subsec:muscle}
\subsubsection{Muscle Exterior Construction Method}
\switchlanguage%
{%
In this study, a three-dimensional muscle exterior was developed by covering a frame made of polypropylene (PP) sheet, which deforms according to wire winding and determines the overall shape, with a braided sleeve (\figref{fig:exterior}).

First, PP sheets are processed as shown in \figref{fig:PP}(a).
The PP sheets are easy to process because a laser machine can cut them.
At the part where the wires are threaded through the PP sheet, a slippery polytetrafluoroethylene (PTFE) tube is inserted to prevent the wires from rubbing against the PP sheet and getting entangled (\figref{fig:PP}(b)).
By arranging multiple arch-shaped materials as a frame, the deformation per unit shrinkage distance is reduced, and a gentle expansion is achieved.
%The deformation of the frame as the muscle contracts is illustrated in Fig. 6, showing how the width increases slightly as the length decreases.
\figref{fig:length} illustrates the frame's deformation as the muscle contracts, showing how the width increases slightly as the length decreases.
The arches are covered with braided sleeves and fixed at both ends so they can be contacted on all surfaces.
}%
{%
  %\hspace{44mm}
  %\break
  
%二次元的な構成での結果から, 三次元的な身体構成と動きの中で使用可能な筋外装には, 表面全体で接触できることと比較的緩慢な膨張変形が必要であった.
本研究では, ワイヤ巻取に応じて変形し全体の形状を決定するポリプロピレン (PP) シートによる骨組みを, 編組スリーブで覆うことで三次元的な筋外装を開発した (\figref{fig:exterior}).
%三次元的な変形筋外装として緩慢な膨張が一つの要件となっていた.
%実際の生物の筋の収縮に伴う断面積変化を測定することは難しいが, 今簡単に体積が一定となると仮定すれば, 収縮距離と幅の関係を表すグラフは下に凸のものとなる.

まずポリプロピレン (PP) シートを\figref{fig:PP}(a)のように加工する.
PPシートはレーザーでカットできるため加工が容易である.
PPシートにワイヤを通す部分では, ワイヤとPPシートが擦れて摩耗したり引っかかったりしないよう滑りやすいポリテトラフルオロエチレン (PTFE) のチューブを挿入する (\figref{fig:PP}(b)).
骨組みとなるアーチ型の材を\figref{fig:length}のように複数配置することで, 単位収縮距離あたりに対する変形を減少させ緩慢な膨張を実現した.
これをbraided sleeveで覆い, 両端を固定することで, 全面で接触できるようにした.
%折り曲げたPPシートを複数束ねて構成した骨組みを編組チューブで覆い, 両端を固定することで, 膨張が緩慢かつ全面で接触可能な筋外装となる (\figref{fig:exterior}).

}%

\begin{figure}[t]
 \centering
 \includegraphics[width=0.7\columnwidth]{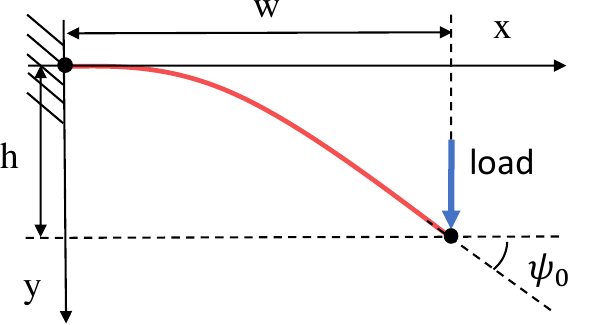}
 %\vspace*{-4.5mm}
  \caption{Flexible bar (based on the figure 2.2 of \cite{1962:frisch:bar})}
  \label{fig:beam}
 %\vspace*{-2mm}
\end{figure}

\subsubsection{Theoretical Calculations of Deformation}

\switchlanguage%
{%
  The deformation of the PP sheet, which constitutes the frame part of the muscle exterior developed in this research, can be modeled as a large deformation of a cantilever beam subjected to a concentrated vertical load at the free end.
  This can be calculated by material mechanics\cite{1962:frisch:bar}.
  In the following, we summarize the beam's deformation for theoretical verification of the deformation of the developed exterior.

The elliptic integrals of the first and second kind, and the complete elliptic integrals of the first and second kind can be written as follows:
\vspace*{-2mm}

\begin{equation}
\vspace*{-1mm}
 %\label{} 第一種楕円積分
 F(\phi,p)=\int^{\phi}_{0} \frac{d\theta}{\sqrt{1-p^{2}\sin^{2}\theta}}
\end{equation}

\begin{equation}
\vspace*{-1mm}
 %\label{} 第ニ種楕円積分
 E(\phi,p)=\int^{\phi}_{0} \sqrt{1-p^{2}\sin^{2}\theta} d\theta
\end{equation}

\begin{equation}
\vspace*{-1mm}
 %\label{} 第一種完全楕円積分
 K(p)=F(\frac{\pi}{2},p)=\int^{\pi/2}_{0} \frac{d\theta}{\sqrt{1-p^{2}\sin^{2}\theta}}
\end{equation}

\begin{equation}
\vspace*{-1mm}
 %\label{} 第ニ種完全楕円積分
 E(p)=E(\frac{\pi}{2},p)=\int^{\pi/2}_{0} \sqrt{1-p^{2}\sin^{2}\theta} d\theta
\end{equation}

In this case, the width $w$ and height $h$ of the beam (\figref{fig:beam}) can be calculated using the beam length $L$ and the shape parameter $p$ (where $1/\sqrt{2} \leq p < 1$) as follows \cite{1962:frisch:bar}:

\begin{equation}
\vspace*{-1.5mm}
 \label{eq:w}
 w=\frac{1}{k}\sqrt{2(2p^{2}-1)}
\end{equation}

\begin{equation}
 \label{eq:h}
 h=\frac{1}{k}\{K(p)-F(\phi_{1}, p)-2E(p)+2E(\phi_{1}, p)\}
\end{equation}

Where $k$ and $\phi_{1}$ are defined as:

\begin{equation}
 %\label{}
 k=\frac{K(p)-F(\phi_{1}, p)}{L}
\end{equation}

\begin{equation}
 %\label{}
 \phi_{1}=\sin^{-1}\left(\frac{1}{\sqrt{2}p}\right)
\end{equation}

Additionally, the shape parameter $p$ is related to the angle $\psi_{0}$ between the beam tip and the horizontal direction as follows:

\begin{equation}
 \sin\psi_{0}=2p^{2}-1
\end{equation}
}%
{%
  %\hspace{40mm}
  %\break
  
  本研究で開発した筋外装の骨組み部分を構成するPPシートの変形は, 自由端で垂直方向に集中荷重がかかる片持梁の大変形としてモデル化できる.
  これは材料力学的に計算可能である\cite{1962:frisch:bar} (\figref{fig:beam}).
  以下では開発した筋外装の変形の理論的な検証に使用するため, 梁の変形についてまとめる.

  今, 第一種楕円積分, 第二種楕円積分, 第一種完全楕円積分, 第二種完全楕円積分はそれぞれ以下のように書ける.

\begin{equation}
 %\label{} 第一種楕円積分
 F(\phi,p)=\int^{\phi}_{0} \frac{d\theta}{\sqrt{1-p^{2}\sin^{2}\theta}}
\end{equation}

\begin{equation}
 %\label{} 第ニ種楕円積分
 E(\phi,p)=\int^{\phi}_{0} \sqrt{1-p^{2}\sin^{2}\theta} d\theta
\end{equation}

\begin{equation}
 %\label{} 第一種完全楕円積分
 K(p)=F(\frac{\pi}{2},p)=\int^{\pi/2}_{0} \frac{d\theta}{\sqrt{1-p^{2}\sin^{2}\theta}}
\end{equation}

\begin{equation}
 %\label{} 第ニ種完全楕円積分
 E(p)=E(\frac{\pi}{2},p)=\int^{\pi/2}_{0} \sqrt{1-p^{2}\sin^{2}\theta} d\theta
\end{equation}

この時, 梁の幅$w$, 高さ$h$は, 梁の長さ$L$, 梁の形状を表す媒介変数$p$により以下のように計算できる \cite{1962:frisch:bar}.

\begin{equation}
 \label{eq:w}
 w=\frac{1}{k}\sqrt{2(2p^{2}-1)}
\end{equation}

\begin{equation}
 \label{eq:h}
 h=\frac{1}{k}\{K(p)-F(\phi_{1}, p)-2E(p)+2E(\phi_{1}, p)\}
\end{equation}

ただし, $k$と$\phi_{1}$はそれぞれ

\begin{equation}
 %\label{}
 k=\frac{K(p)-F(\phi_{1}, p)}{L}
\end{equation}

\begin{equation}
 %\label{}
 \phi_{1}=\sin^{-1}\left(\frac{1}{\sqrt{2}p}\right)
\end{equation}

である. また梁の形状を表す媒介変数pは, 梁先端と水平方向がなす角度$\psi_{0}$について以下の関係がある.
\begin{equation}
 \sin\psi_{0}=2p^{2}-1
\end{equation}

}%

\begin{figure}[t]
 \centering
 \includegraphics[width=1.0\columnwidth]{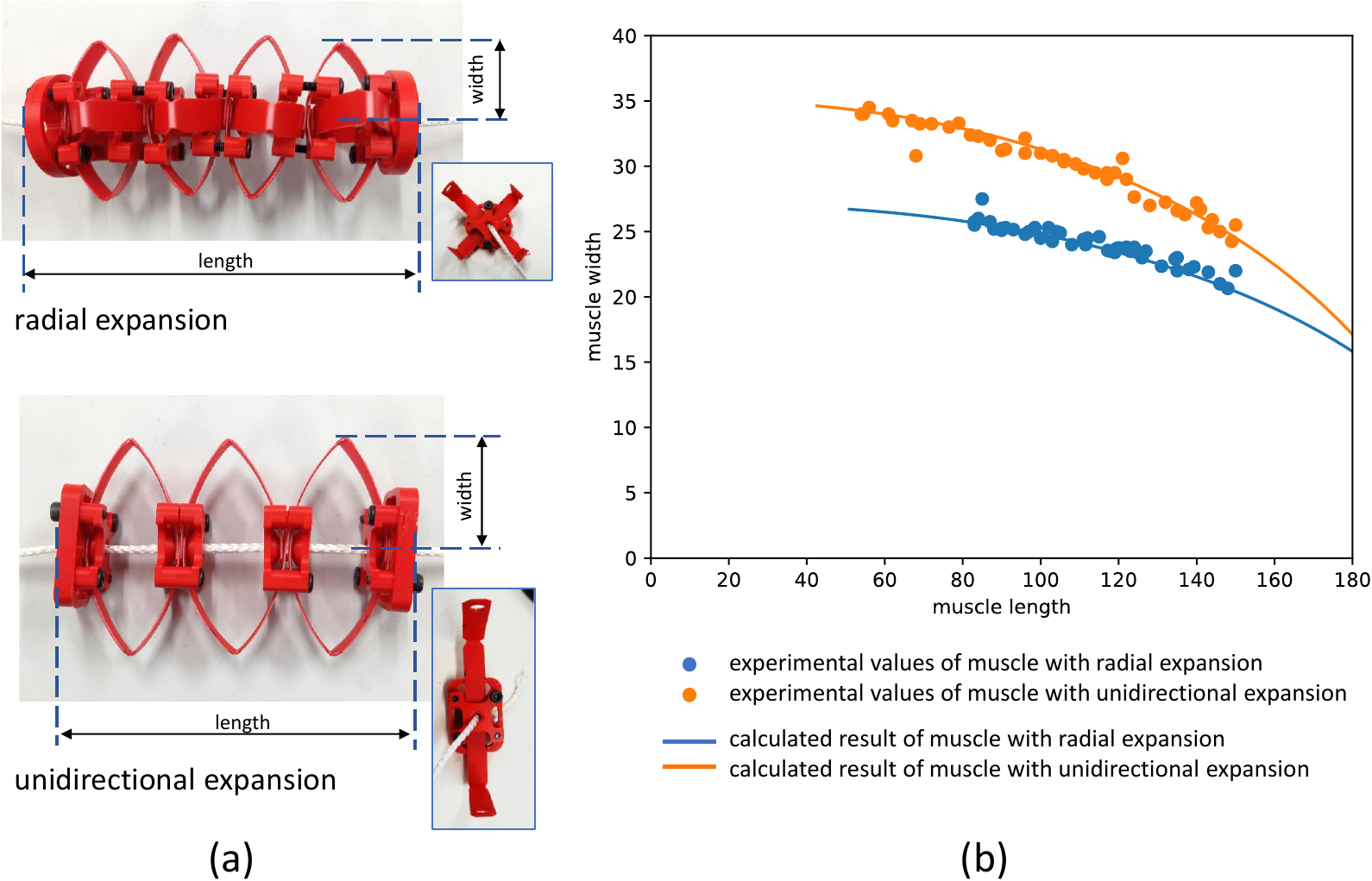}
 %\vspace*{-6mm}
  \caption{(a) Two types of frames made of PP sheets. (b) Lines show calculated values, plots show experimental values. Calculated result is representing the experimental result well.}
  \label{fig:deform}
 %\vspace*{-4mm}
\end{figure}

\subsubsection{Deformation of Muscle Exteriors}

\switchlanguage%
{%
%In this study, we have developed two types of exterior: one expands isotropically in the direction perpendicular to the contraction, and the other expands planarly.
In this study, we have developed two types of exterior: one achieves radial expansion in the plane perpendicular to the contraction, and the other exhibits unidirectional expansion perpendicular to the contraction (referred to as planar muscle in this study).
We verify the measured results of actual deformations with theoretical calculations for these two types of muscles.
The length shown in \figref{fig:PP}(a) corresponds to the length $L$ of the beam.
The theoretical calculations of  $w$ and $h$ correspond to the width and length shown in \figref{fig:exterior}(a).

The difference in shape is realized by changing the direction of the arch structure in the frame, four directions in the former and two directions in the latter (\figref{fig:deform}(a)).
Let $n$ be the number of arches in the length direction and $h_0$ be the offset due to sheet thickness and components, then the width and length of the muscle exterior are multiplied using Eq.~\eqref{eq:w} and Eq.~\eqref{eq:h} as follows:
\begin{equation}
 %\label{}
 (width)=w(L, p)
\end{equation}
\begin{equation}
 %\label{}
 (length)=n \times h(L, p)+h_0
\end{equation}
For the muscles with radial expansion, $n$=8, $L$=\SI{27}{\mm}, $h_0$=\SI{22}{\mm}.
For the planar muscles, $n$=6, $L$=\SI{35}{\mm}, $h_0$=\SI{14.5}{\mm}.
These values were determined experimentally, mainly from the size of the braided sleeves.

The actual measured and theoretically calculated values for the length-width relationship of these frames are shown in \figref{fig:deform}(b).
The results confirm that the calculated deformations of the cantilever beams are well representative of the actual deformations.
The two types implemented in this research have approximately the same volume, and the plane muscles are more expansive.
It can be said that the overall deformation can be designed with the size and number of deformed beams as parameters.
}%
{%
  %\hspace{30mm}
  %\break

本研究では, 膨張が収縮に直交する方向に当方的な筋と, 面状の筋の2種を開発した.
これらについて, 実際の変形の測定結果を理論的な計算により検証する.
\figref{fig:PP}(a)に示した長さは, 梁の長さ$L$に相当する.
%形状によらず, 本研究の手法においては共通して以下の過程を経る.
%ここで梁の長さに相当する値は設計する変形形状に応じて変化する.

骨組みにおけるアーチ構造の出る方向を変えることで形状の違いを実現しており, 前者では4方向, 後者では2方向となっている (\figref{fig:deform}(a)).
アーチ構造の筋長さ方向での数を$n$, シートの厚みなどによる筋の長さ方向へのオフセットを$h_0$とすると, 筋のwidthとlengthは\eqref{eq:w, eq:h}を用いて以下のようにかける.
\begin{equation}
 %\label{}
 (width)=w(L, p)
\end{equation}
\begin{equation}
 %\label{}
 (length)=nh(L, p)+h_0
\end{equation}
膨張が当方的な筋では$n=8$, $L$=\SI{27}{\mm}, $h_0$=\SI{22}{\mm}である.
面状の筋では$n=6$, $L$=\SI{35}{\mm}, $h_0$=\SI{14.5}{\mm}である.
これらの値は主に編組スリーブのサイズから実験的に決定された.

実際にこれらの骨組みの長さと幅の関係について, 測定値と理論的に計算される値を\figref{fig:deform}(b)に示す.
この結果から片持梁の変形計算は実際の変形をよく表していることが確認された.
今回実装した2種類は, だいたい同程度の体積としており, 面状の筋がより膨らむものとなっている.
変形する梁のサイズと個数をパラメータとして全体の変形を設計することができると言える.
}%

\subsection{Tendon based on SAT} \label{sec:tendon}
\subsubsection{Tendon Construction Method and Mechanical Characteristics}
\begin{figure}[t]
  \centering
      \includegraphics[width=1.0\columnwidth]{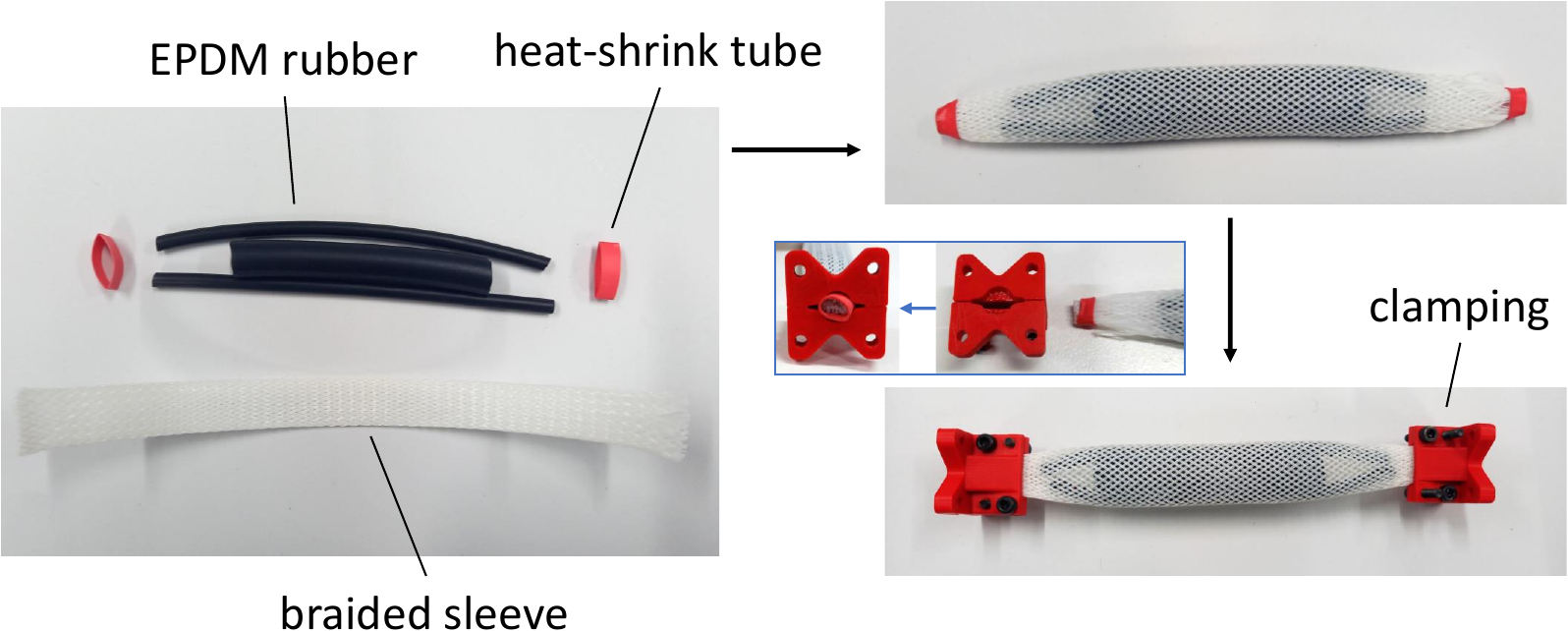}
 \vspace*{-2mm}
  \caption{Assembly of tendon based on SAT\cite{robomech2003:shirai:sat}.}
  \label{fig:sat}
\end{figure}

\begin{figure}[t]
  \centering
      \includegraphics[width=1.0\columnwidth]{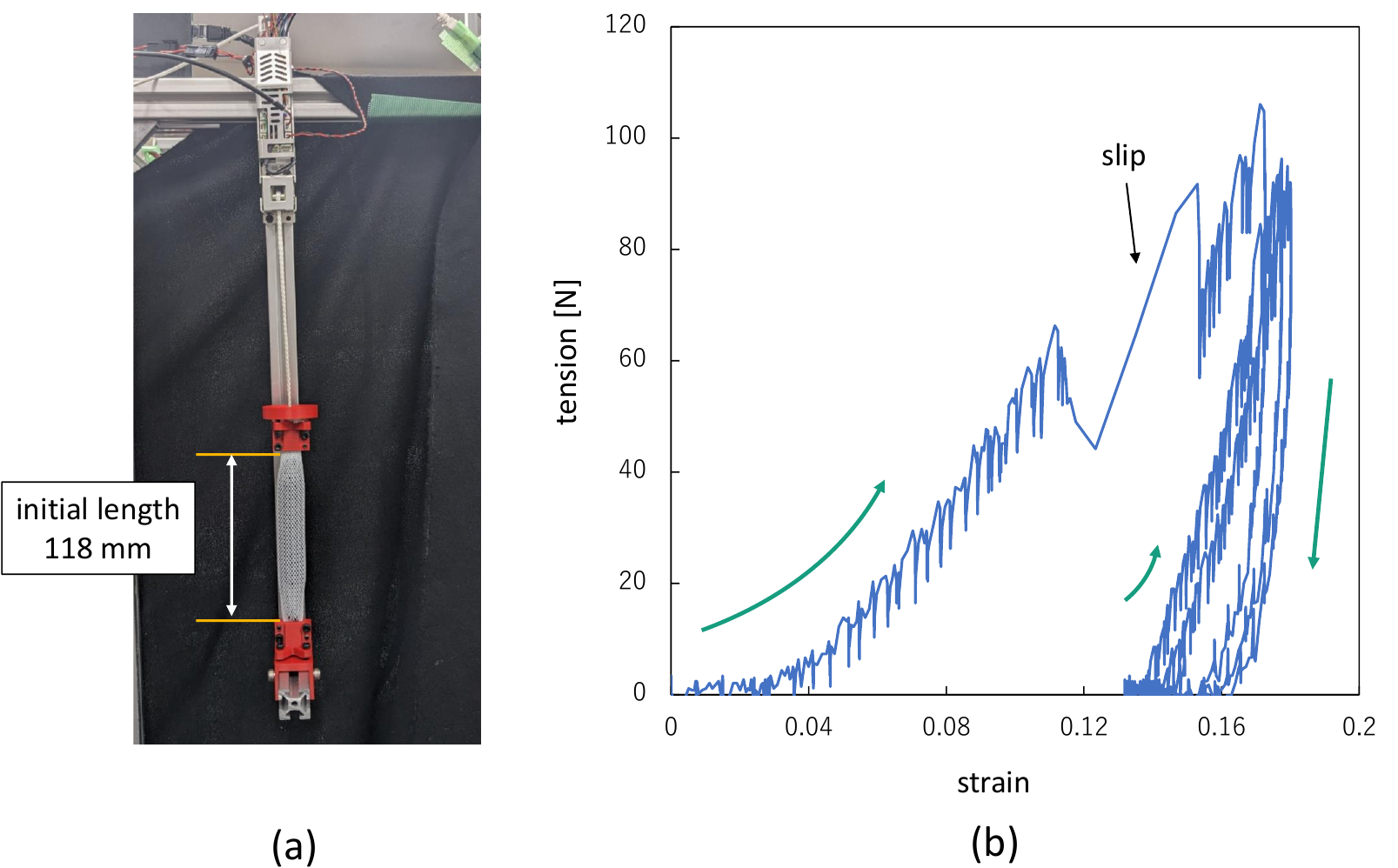}
  \vspace*{-2mm}
  \caption{(a) Tensile experiment of tendon based on SAT. (b) The elongation was large when the load was first applied. Slipping also occurred, but it did not come out of the fastener. After that, the strain remained within a narrow range.}
  \label{fig:tensile}
\end{figure}

\switchlanguage%
{%
As a tendon that can be used in a three-dimensional configuration, it should be thin and strong, with low friction on the surface and no interference during contact.
It is also known that the stiffness can be changed by incorporating a nonlinear elastic element in the wire drive\cite{AR1997:koganezawa:antagonistic}, which is important for expanding the range of robot motions.
Stiffness Adjustable Tendon (SAT)\cite{robomech2003:shirai:sat} has already been proposed as a nonlinear elastic element with a tendon-like shape.
By covering a silicon sponge with a braided tube, the braided sleeve is pulled and crushes the sponge's inside, resulting in nonlinear elasticity as a whole.
This method is considered to be durable and can withstand contact at joints because the surface gap is smaller than that of conventional springs.

Therefore, in this study, we constructed an elastic element as a tendon based on the proposed SAT (\figref{fig:sat}).
In order to make the element not only cylindrical in shape, several round rubbers are inserted inside and the ends of the sleeves are fastened in a plane shape.

Ethylene-propylene-diene (EPDM) rubber was used as the inner material.
To keep the rubber from coming out, the end of the braided sleeve is covered with heat-shrinkable tubing and then heated to close it.
Since the heated part becomes thicker, it should protrude outside the fastener during fastening to prevent it from falling out.

The relationship between load and strain of the tendon based on the SAT was examined when the vertical upward load was increased or decreased in steps by discretely changing the command current value to the motor (\figref{fig:tensile}).
%Although the fastener sometimes slipped during the experiment, it did not come out of the fastener because the endpoint was thicker than the fastener's hole.
%Although the elongation was large at the first increase in load due to the margin of the sleeve's shape, the strain was kept within a small range thereafter.
The fastener occasionally slipped during the experiment but did not detach completely, as the endpoint was thicker than the fastener's hole.
Although the elongation was large at the initial load increase due to the margin of the sleeve's shape, the strain was kept within a small range after that.
The tendon, based on the SAT, has the required thinness and strength. The entire tendon is covered by the braided sleeve, which is thought to prevent interference even when the tendon contacts joints.
}%
{%
三次元的構成で使用できる腱として, 細さと強さを両立し, 表面での摩擦が小さく接触時に干渉しないものが望ましい.
%二次元的構成では光造形したゴムを用いていた (\figref{fig:2D}) が, サイズに対して弱く, 劣化や破断が起きていた.
また, ワイヤ駆動に非線形弾性要素を取り入れることで剛性を変化させられるということが知られており\cite{AR1997:koganezawa:antagonistic}, ロボットの動作の幅を広げる上で重要である.
腱のような形状の非線形弾性要素としては, 既にStiffness Adjustable Tendon (SAT) \cite{robomech2003:shirai:sat}が提案されている.
シリコンスポンジを編組チューブで覆うことで, 引っ張られたbraided sleeveが中のスポンジを押しつぶすという現象が生じて, 全体として非線形な弾性を示す.
この手法は, 耐久性に優れ, 一般的なバネと比較して表面の隙間が小さいことから関節部での接触に耐えうると考えられる.

そこで本研究では, 提案されているSATを元に, 腱となる弾性要素を構成した (\figref{fig:sat}).
円柱形に限らないものとするため, 内部に複数の丸ゴムを入れスリーブの端を面状に締結した.
%
%差分は以下の3点である.
%\begin{itemize}
% \item 内部に入れる丸棒をスポンジからより硬いゴムに変更した
% \item 内部に複数の丸棒を入れることで単なる円柱形に限らない形状を可能とした
% \item 端点を面状に締結することで骨格に沿うようにした
%\end{itemize}
%
内部の材としてはエチレンプロピレンジエン (EPDM) ゴムを使用した.
ゴムが外に出ないよう, 編組スリーブの端を熱収縮チューブで覆った上で熱することで閉じるようになっている.
熱した部分が太くなるので, そこが締結時には締結具の外にはみ出るようにすることで締結具から抜けることを防ぐ.

このSATを元にした腱について垂直上方への荷重を段階的に増減させた際の荷重とひずみの関係を調べた (\figref{fig:tensile}).
実験中に締結部が滑ることがあったが, 端点部分が締結具の穴より太いため外れることはなかった.
1回目の荷重増加ではsleeveの形状の余裕などにより伸びが大きいものの, その後はひずみが小さい範囲内に収まっている.
%このグラフから先に開発した, より太いゴム腱と比較して硬さが大きいことが分かる.
%また荷重の増加過程において非線形性が見られる.
グラフからこの腱要素が強く, 非線形性を示すことが読み取れる.
SATを元にした腱は, 要件としていた細さと強さを備え, 全体が編組スリーブで覆われていることで関節での接触でも干渉しないと考えられる.
}

\begin{figure}[t]
  \centering
      \includegraphics[width=1.0\columnwidth]{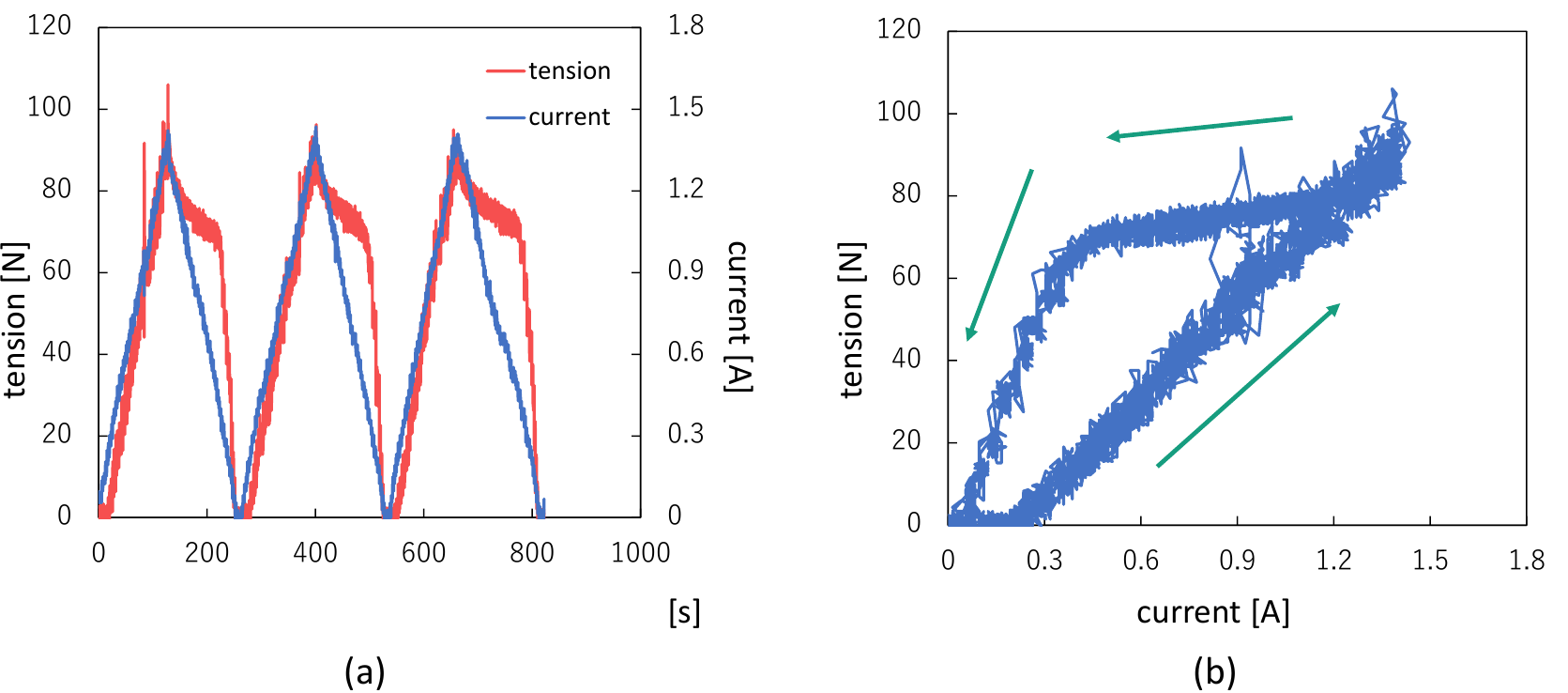}
  \vspace*{-2mm}
  \caption{(a) Time series of tension and current during tensile experiment. (b) Wire winding module has hysteresis in tension-current characteristics.}
  \vspace*{-2mm}
  \label{fig:add-tensile}
\end{figure}

\subsubsection{Tension-Current Characteristics of Wire Winding Module}
\switchlanguage%
{%
In relation to the tensile experiments, we also report on the characteristics of the wire winding module by examining the relationship between the tension and the current.
The time-series variation of tension and current is shown in \figref{fig:add-tensile}(a), and the tension-current graph is shown in (b).

Ideally, the motor's exerted torque and the resulting wire tension are proportional to the motor current.
When the current is increased, they are almost proportional to each other.
However, when the current is decreased, the tension value does not decrease much, and a sharp decrease is observed from a certain value.
Also, the tension value does not initially increase when the current increases.

This result is a characteristic of the wire winding module itself, and it can be said that a force resisting the direction of motion of the wire and the pulley is working.
This is considered to be caused by the reduction gear in particular, and the effect is expected to be reduced by decreasing the reduction ratio.

%Although not treated in this research, the effect of hysteresis in the tension-current relationship, including control, must be considered when considering biological-like motion.
Although not addressed in this research, the effect of hysteresis in the tension-current relationship must be considered for biological-like motion, including its impact on control.
Hysteresis is expected to negatively affect fast movements, in which case the motor module needs to be improved.
}%
{%
また引張実験に関連して, 張力と電流の関係を調べることでワイヤ巻取モジュールの性質についても報告する.
張力と電流の時系列変化を\figref{fig:add-tensile}(a)に, 張力-電流グラフを(b)に示す.

理想的にはモータの発揮トルクとそれによるワイヤ張力はモータ電流に比例する.
実際電流を増加させた際, これらはほとんど比例関係にある.
しかし電流を低下させた際には張力値があまり減少せず, ある値から急激に減少する様子が観察される.
%このヒステリシス特性は同種のモータモジュールとワイヤを用いて錘を上下させる実験でも報告されており, 腱ではなくモータモジュール自体の特性である.
% \cite{2016:ookubo:master}
また電流を増加させる時についても, はじめの内は張力値が増加していない.

この結果はワイヤ巻取モジュール自体の特性であり, ワイヤとプーリの運動方向に抵抗する力が働いていると言える.
これは特に減速機に起因すると考えられ, 減速比を小さくすることで影響を低減できると予想される.

本研究では扱わないが, 生物らしい動きを考える上では制御も含めて張力-電流の関係のヒステリシスがどのような影響を与えるか注視する必要がある.
素早い動きを行う上ではヒステリシスが悪影響を及ぼすと予想でき, その場合モータモジュールの改善が必要となる.
}%

\begin{figure*}[tbh]
  \centering
      \includegraphics[width=2.0\columnwidth]{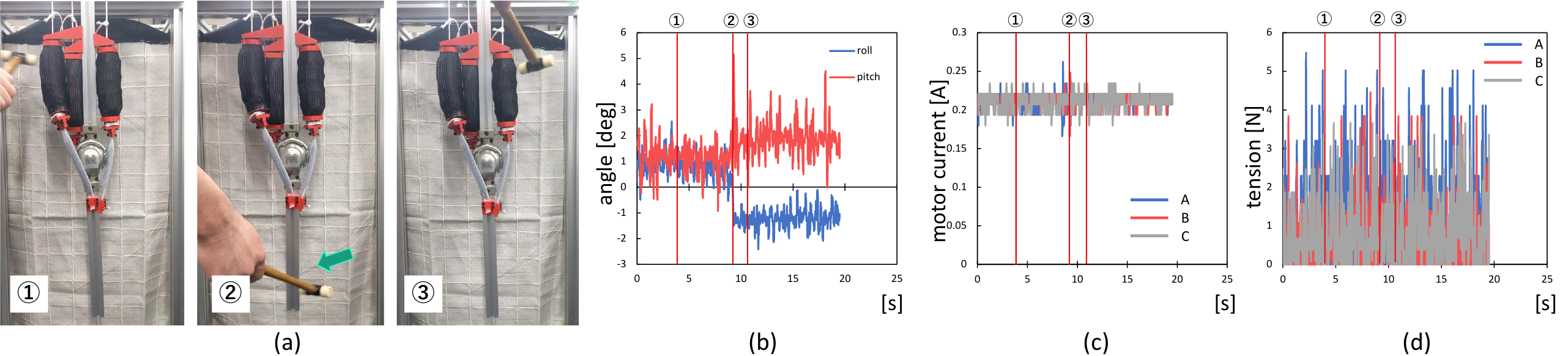}
  \caption{The structure protects inner parts against hammering. (a) shows the scene of the hammer impact. (b)(c)(d) shows the time series of measured joint angle, motor current, and tension. The command current is constant.}
  %\vspace*{-4mm}
  \label{fig:6-hammer}
\end{figure*}

\begin{figure*}[tbh]
  \centering
      \includegraphics[width=2.0\columnwidth]{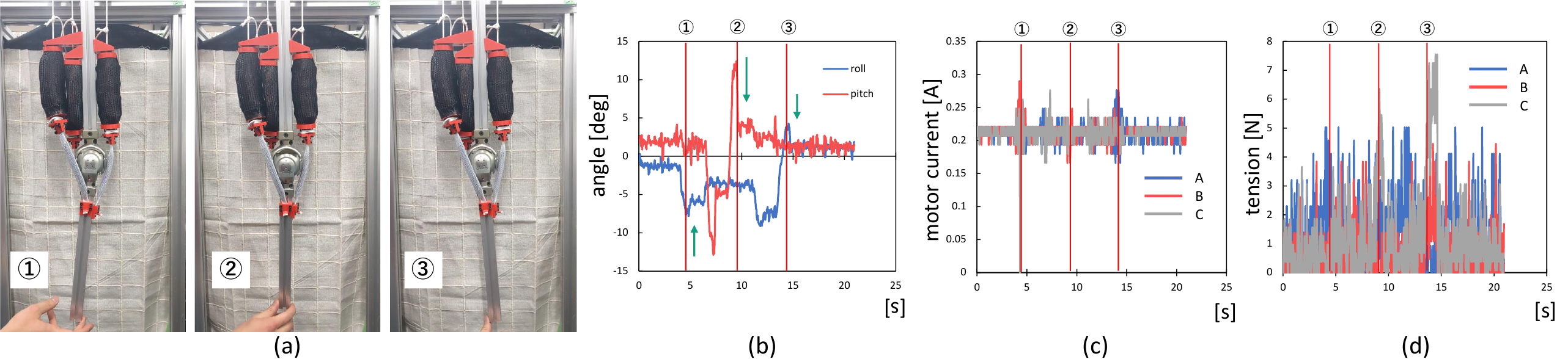}
  \caption{2-axis 3-muscle robot with low tension. (a) shows the scene before releasing the arm. (b)(c)(d) shows the time series of measured joint angle, motor current, and tension. The command current is constant. The arm part moves softly and then converges.}
  \label{fig:6-low}
\end{figure*}

\begin{figure*}[tbh]
  \centering
      \includegraphics[width=2.0\columnwidth]{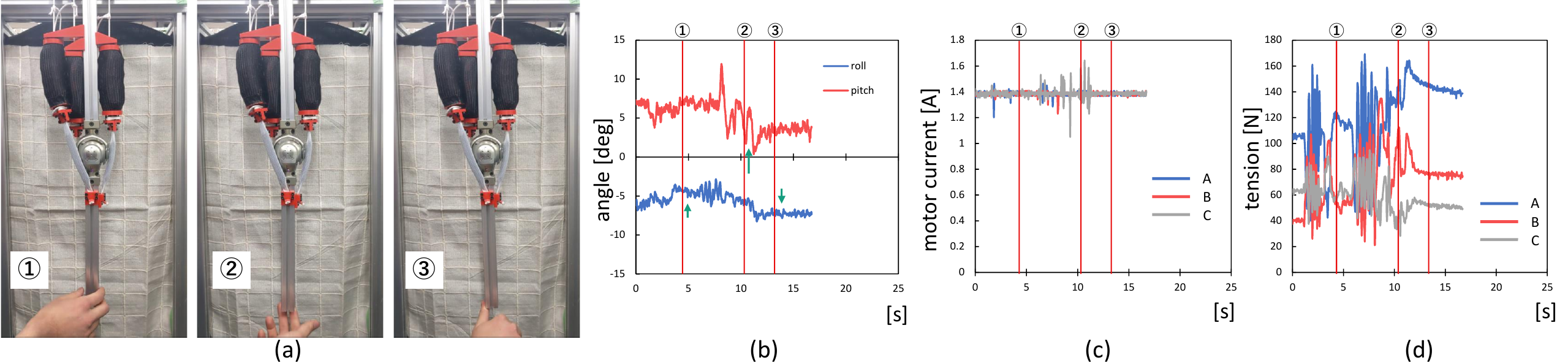}
  \caption{2-axis 3-muscle robot with high tension. (a) shows the scene before releasing the arm. (b)(c)(d) shows the time series of measured joint angle, motor current, and tension. The command current is constant. The arm part is almost motionless and converges immediately.}
  \label{fig:6-high}
\end{figure*}

\begin{figure}[tbh]
  \centering
      \includegraphics[width=1.0\columnwidth]{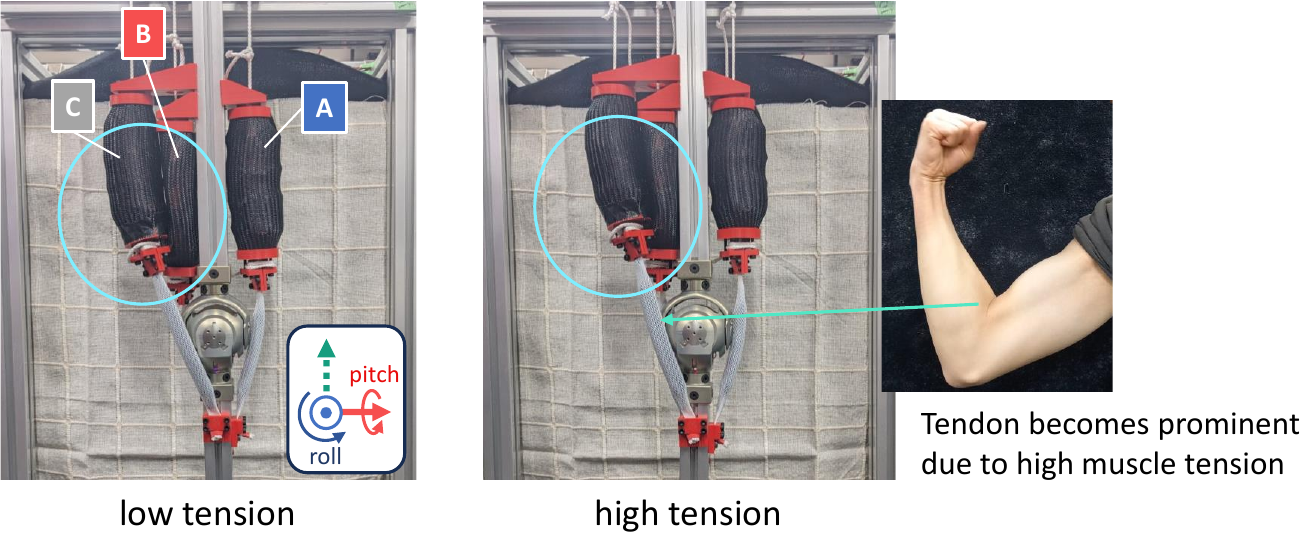}
  \caption{When muscle tension is high, the muscle-tendon complex is pressed inward (the area indicated by the light blue circles). Also, the tendon becomes stretched and more visible.}
  \label{fig:6-app}
\end{figure}

\begin{figure*}[tbh]
 \centering
  \includegraphics[width=2.0\columnwidth]{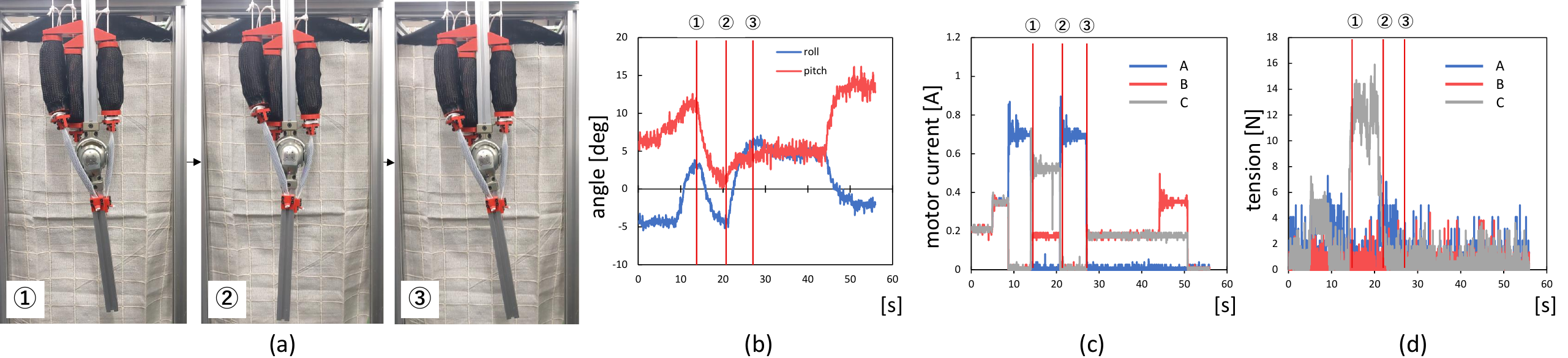}
  %\vspace*{-4mm}
  \caption{Arm moving experiment without disturbances. (a) The robot arm can move in three-dimensional space. (b)(c)(d) shows the time series of measured joint angle, motor current, and tension. By changing the current command value, the tension on the wire changes, and the joint moves.}
  \label{fig:6-move}
\end{figure*}

\begin{figure*}[tbh]
 \centering
  \includegraphics[width=2.0\columnwidth]{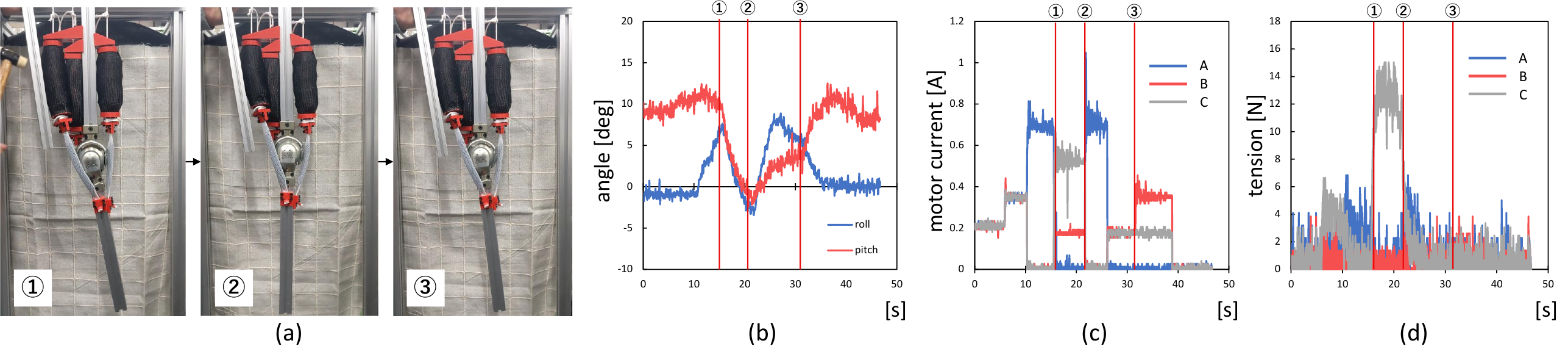}
  %\vspace*{-4mm}
  \caption{(a) The robot maintains moving in spite of external disturbance. (b)(c)(d) shows the time series of measured joint angle, contraction length, and tension. By changing the current command value, the tension on the wire changes, and the joint moves. The trajectory of joint angles shows no sudden or unexpected changes during the disturbance experiments. The observed variations in the angle measurements remain consistent with the noise level seen in the undisturbed motion.}
  \label{fig:contact}
\end{figure*}

\begin{figure}[tbh]
  \centering
      \includegraphics[width=1.0\columnwidth]{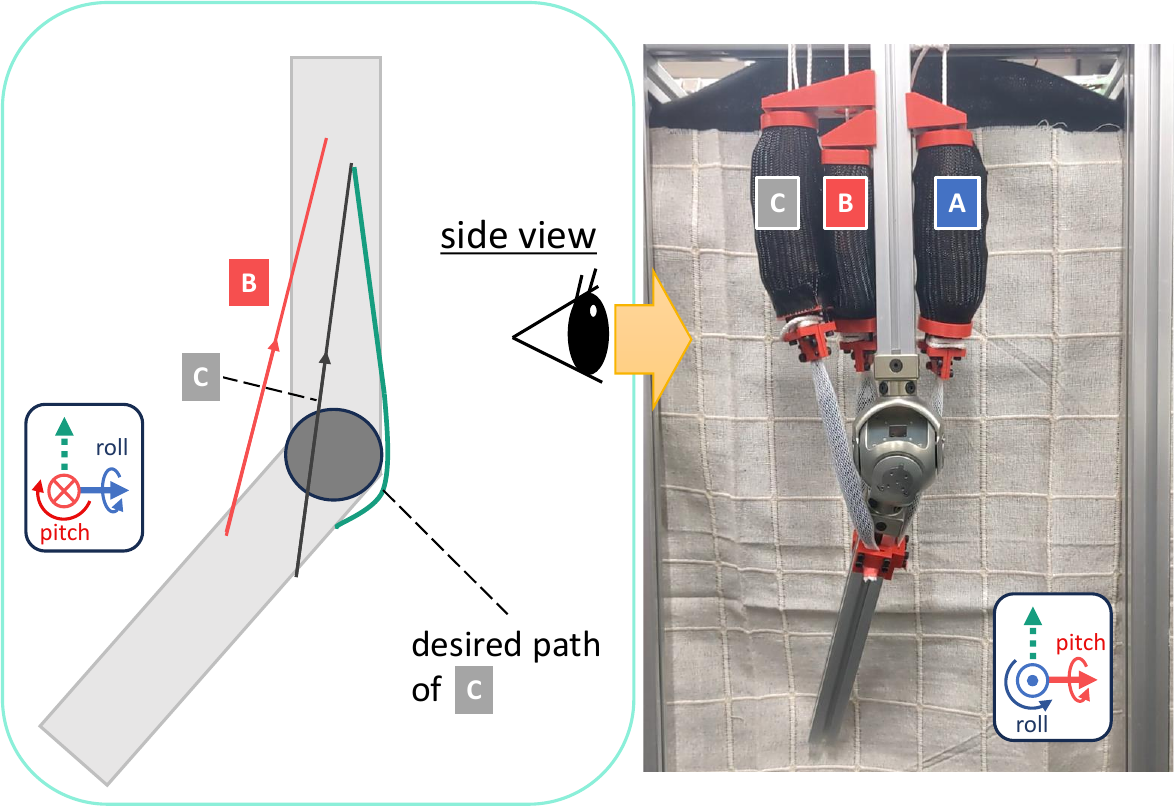}
  \caption{The tendon path sometimes becomes dislocated from the joint.}
  \vspace*{-4mm}
  \label{fig:path}
\end{figure}

\section{Experiments and Discussions} \label{sec:experiments}
\switchlanguage%
{%
In this chapter, we evaluate whether the advantages of the muscle-tendon complex drive concept discussed and verified in the two-dimensional type of ww-MTC\cite{MTC-2D:Ribayashi:HUMANOIDS2023} are valid in three-dimensional motion, and whether the configuration proposed in this research is also effective in three-dimensional motion.
Specifically, the following experiments will be conducted to verify the characteristics of the robot with 3D ww-MTC.
\begin{itemize}
\item We verify the robot's toughness (internal protection function) by applying disturbance with a hammer.
\item We verify whether the muscles cooperate without interfering with each other by examining the change in stiffness when the tension is changed. We also examine the change in the appearance of the muscles in response to changes in tension.
\item We examine the adaptability of this robot to contact with the environment by applying disturbance while moving the arm part.
\end{itemize}
}%
{%
本章ではtwo-dimensional type of ww-MTC\cite{MTC-2D:Ribayashi:HUMANOIDS2023}にて議論・検証された筋腱複合体駆動コンセプトの利点が3次元的な動きにあっても成立するか, 本研究で提案した構成が3次元的な動きにおいても有効かを評価する.
%本章ではtwo-dimensional type of ww-MTC\cite{MTC-2D:Ribayashi:HUMANOIDS2023}にて議論・検証された筋腱複合体駆動コンセプトの利点が, 3次元的な動きにあっても成立するか, 3次元的な動きにおいても有効な駆動方式として筋腱複合体駆動方式を構成できたかを評価する.
具体的には以下の実験により3次元的ww-MTCを用いたロボットの性質を検証する.
\begin{itemize}
\item ハンマーで外乱を加えることで, 頑健性 (内部保護機能) を検証する.
\item 張力を変えた際の剛性の変化を調べることで, 筋同士が互いに阻害せず協調できているかを検証する. また張力変化に応じた見た目の変化について考察する.
\item アーム部を動かしながら外乱を加えることで, このロボットの環境接触への適応性を検証する.
\end{itemize}
}
\subsection{2-axis 3-muscle Robot Configuration} \label{subsec:robot}
\switchlanguage%
{%
The robot used in the experiment is shown in \figref{fig:abst}(b).
The joint modules used in Musashi\cite{iros2019:kawaharazuka:musashi} are used for the 2-axis joints.
The module has a potentiometer inside to measure the angles.
Since the actual value of the potentiometer is noisy, the joint angle graph in \secref{subsec:result} is a calculated moving average.

The wire winding module is mounted far from the muscle.
The ability to separate the drive from the actual force-activated part is an advantage of the wire drive.
In this case, the space is open enough to connect the wires on the muscle exterior and the wires from the motor module, facilitating assembly and repair in case of wire breakage.

In mounting the ww-MTC, the tendons are adjusted to align around the joint without loosening when the roll and pitch are 0 degrees, and the muscle is around its natural length with no tension applied to the wires.
Three ww-MTCs are attached to the two axes, and the tendon ends of two muscles (B, C) on the same side of the central link are attached at different positions to allow three-dimensional arm movement.
In addition, a planar muscle is used for the medial muscle (B in \figref{fig:abst}(b)) in the area where the muscles overlap.
This reduces the effect of the inner muscles pushing the outer muscles during deformation so that the muscles can cooperate effectively. This corresponds to the relationship between the brachialis and biceps muscles in the human body\cite{book2009:tank:anatomy}, for example.
}%
{%
実験に用いたロボットは\figref{fig:abst}(b)の通りである.
%三次元的な構成と動きにおける提案手法の有効性を示すため, ロール, ピッチの2軸を有する3筋のロボットを筋腱複合体により構成した(\figref{fig:abst}(b)).
2軸の関節部にはMusashi\cite{iros2019:kawaharazuka:musashi}で使用された関節モジュールを用いた.
内部にポテンショメータが内蔵されており, 角度を測定することが可能である.
ポテンショメータの実際の値はノイズが大きいため, \secref{subsec:result}における関節角度グラフは移動平均を計算したものとなっている.

ワイヤ巻取りモジュールは筋から離れた位置に搭載されている.
駆動部と実際に力が作用する部分を切り離せることはワイヤ駆動の利点である.
ここでは筋外装にあるワイヤとモータモジュールからのワイヤを結ぶことができるだけのスペースを開けることで, 組立とワイヤ破断時の修理が容易となっている.

ww-MTCの取付に当たっては, ロール・ピッチが0度, 筋がワイヤに張力をかけない状態での自然な長さ前後である時に腱が関節周りに緩まずに沿うよう調整した.
2軸に対して3つのww-MTCが取り付けられており, アーム部を3次元的に動かすことができるよう中心のリンクに対して同じ側にある2つの筋 (B, C) について腱端が前後の異なる部分につけられている.
また, 筋が重なる部分の内側 (\figref{fig:abst}(b)におけるB) の筋について, 外側の筋の動きを阻害しないよう面状の筋を用いた.
%当方的な膨張と面状の膨張を設計したが, 2つの筋が重なる部分の内側 (B) を面状とし, 他の2つは当方的な膨張を示すものとした.
これは変形時に内側の筋が外側の筋を押しやる効果を低減することで筋同士が有効に協調できるようにするものであり, 人体では例えば上腕筋と上腕二頭筋の関係\cite{book2009:tank:anatomy}に対応する.
}%
\subsection{Internal Protection} \label{subsec:result}
\switchlanguage%
{%
Muscle-tendon complexes are considered to have toughness in protecting themselves and the body when in contact with the environment.
We conducted an experiment in which we hit each part of a 2-axis 3-muscle robot with a hammer (\figref{fig:6-hammer}).
There was no damage to ww-MTC, internal parts, or wires, and the toughness of the robot was confirmed.

In addition, a two-dimensional configuration using a muscle exterior with a single arch structure (\figref{fig:2D}) showed a rapid increase in tension when impact was applied\cite{MTC-2D:Ribayashi:HUMANOIDS2023}, but this was not observed in the proposed configuration.
Compared to the structure using only two PTFE plates, the muscle exterior composed of multiple PP sheets absorbs external forces without transferring them to the internal wires because of the structural margin that allows for distortion and deformation.
This result indicates the high flexibility and safety of the proposed muscle exterior.
}%
{%

環境接触において筋腱複合体は体内やそれ自体を保護する頑健性を有すると考えられる.
2軸3筋構成のww-MTCロボットにおいてもハンマーで各部を叩く実験を行った (\figref{fig:6-hammer}).
ww-MTCや内部パーツ, ワイヤ等にも損傷はなく, 頑健性が確認された.

また, 1つのアーチ構造による筋外装を用いた二次元的な構成 (\figref{fig:2D}) では衝撃が加わった際に急激な張力上昇が見られたが\cite{MTC-2D:Ribayashi:HUMANOIDS2023}, 今回の構成では観察されなかった.
%これは外装部分が複数の構造材で構成されることで, 二次元的な構成で実現した筋外装や空気圧人工筋肉と異なり, 接触による力や形状変化が接触部のみで吸収され他の箇所と外装全体への影響が小さくなることによる効果だと考えられる.
このことから2枚のPTFE板のみを用いる構造と比較し, 複数枚のPPシートを組み合わせた筋外装は, 構造上の余裕があり歪みや変形を許容できることで, 外力を内部のワイヤに伝えることなく吸収していると考えられる.
これは提案した筋外装の高い柔軟性と安全性を示すものである.
%外装部分が衝撃を効果的に吸収していると結論づけられる.
}

\subsection{Stiffness and Appearance when Changing Tension}
\switchlanguage%
{%
In order to verify the changes in stiffness and appearance, experiments were conducted to move the tip of the arm in various directions and then release it for the cases of low (\figref{fig:6-low}) and high (\figref{fig:6-high}) motor command current.
In the high-tension case, moving the arm by hand force was difficult.
Especially in the direction of roll, where the moment arm of the muscle is large, the arm hardly moves.
In the two-dimensional ww-MTC\cite{MTC-2D:Ribayashi:HUMANOIDS2023}, the muscles expanded too much and interfered with each other's movement, making it difficult to move even at low tension, and no change in stiffness was observed.
On the other hand, in the present work, the change in the angle of the arm after the release indicates that the arm moves softly at low tension, while the stiffness increases at high tension.
These results show that the muscle expansion is gentle enough to allow the muscles to cooperate effectively.

In terms of appearance, the muscular appearance of the robot is useful for coexistence with humans by making the robot's state more visible.
Comparing the still images in the two states, we can see that the lower end of the muscle is higher in the case of high tension (\figref{fig:6-app}).
In the higher tension case, the outer muscle (C) presses down on the inner muscle (B), and the tendon is more nearly parallel to the body axis.
%Although we cannot clearly see the expansion compared to the two-dimensional ww-MTC constructed with PTFE (\figref{fig:2D}(b)), we can see the difference from the elongation of the tendons.
Although the expansion is not as visually apparent as in the two-dimensional ww-MTC constructed with PTFE (\figref{fig:2D}(b)), we can observe the difference through the elongation of the tendons.
These differences may appear as wrinkles or shadows when the body structure, including the skin, is constructed, and further verification is needed.
}%
{%
剛性と見た目の変化について検証するため, モータ指令電流が低い場合 (\figref{fig:6-low}) と高い場合 (\figref{fig:6-high}) について, アーム部先端を様々な方向に動かしてからリリースする実験を行った.
張力が高い場合では, 手の力でアームを動かすこと自体が困難であった.
特に筋のモーメントアームが大きいroll方向にはほとんど動かなかった.
二次元的ww-MTC\cite{MTC-2D:Ribayashi:HUMANOIDS2023}では筋同士が膨張しすぎて互いの動きを邪魔し, 低い張力でも動きにくかったため, 剛性の変化が見られなかった.
一方で本研究ではリリース後のアームの角度変化から, 低い張力では柔らかく動き, 高い張力では剛性が上がったと言える.
このことから, 筋の膨張は筋同士が有効に協調動作できる程度にゆるやかであると言える.
%張力が低い場合でもリリース後すぐに収束するが, そこまでにある程度の角度動くことを考えると2つの状態で剛性が異なると言える.

見た目について, 筋肉質な見た目はロボットの状態を視認しやすくすることで人間との共存に役立つ.
2つの状態での静止画を比較すると, 張力が高い場合の方が筋の下端が高い位置にあることが分かる (\figref{fig:6-app}).
高い張力では外側の筋 (C) が内側の筋 (B) を押さえつけ, 腱がより身体の軸に対して並行に近くなっている.
PTFEで構成した二次元的ww-MTC(\figref{fig:2D}(b))と比較してはっきりと膨張を見て取ることはできないが, 腱の伸びから違いを読み取ることは可能である.
これは皮膚を含めた身体構造を構成した際に明確な皺や影となって表れる可能性があると考えられ, さらなる検証が必要である.
}

\subsection{Adaptability to Environmental Contact during Arm Movement}
\switchlanguage%
{%
Motion experiments of a 2-axis 3-muscle robot were conducted, in which the arm part is swung.

The motion experiment without any disturbance is shown in the \figref{fig:6-move}, and the muscles move while maintaining contact.
On the other hand, a similar behavior was observed when the muscles were subjected to external disturbances such as pressing a square aluminum rod against them, shaking them with a surface pressed against them, or hitting them with a hammer (\figref{fig:contact}).
The trajectory of joint angles does not show sudden or unexpected changes (\figref{fig:contact}(b)).
Although the external disturbance changed the position and shape of the muscles, they could continue to move while maintaining contact with each other.

As in the case of \figref{fig:6-hammer}, no significant change in tension was observed, and both the muscles and tendons could continue to be used without any problems after the experiment.
It is challenging to conduct disturbance experiments safely with standard wire drives because of the possibility of snagging and a sudden increase in tension.
%It is challenging to conduct safe experiments with standard wire drives because of the possibility of snagging and a sudden increase in tension.
It is considered that the flexible deformable muscle exterior, which can be contacted on all surfaces, has enhanced the adaptability to environmental contact.

%On the other hand, depending on the trajectory, the tendon's path may drop out of the joint, as shown in \figref{fig:path}, making it difficult to move in a certain direction.
On the other hand, depending on the trajectory, the tendon's path may dislocate from the joint, as shown in \figref{fig:path}, making it difficult to move in a certain direction.
%Although we used a pseudo spherical joint module\cite{iros2019:kawaharazuka:musashi} in this study, when using muscle-tendon complex drive in a three-dimensional configuration, it is necessary to design the structure of the joint part as well, taking the range of motion into consideration.
We used a joint module\cite{iros2019:kawaharazuka:musashi} that provides a wide range of motion similar to a spherical joint in a compact form.
However, when using muscle-tendon complex drive in a three-dimensional configuration, it is necessary to design the structure of the joint part as well, taking the range of motion into consideration.
}%
{%
2軸3筋ロボットについてアーム部を振るような動作実験を行った.

外乱の無い場合での動作実験は\figref{fig:6-move}のようになり, 筋同士が接触を保ちながら動作している.
一方筋に対してアルミ角材を押し付ける, 面を押し当てた状態で揺する, ハンマーで叩くなどの外乱を加えた場合でも同様に動作することができた (\figref{fig:contact}).
外乱によって筋の位置や形状は変化しながらも, 筋同士が接触を保ちながら動作を継続することが可能であったと言える.

\figref{fig:6-hammer}と同様張力に大きな変化は見られず, 実験後は筋・腱ともに元の状態で問題なく使用を継続することができた.
一般的なワイヤ駆動では引っ掛かりや張力の急激な上昇などが起こる可能性があり, 安全に実験を行うことも難しい.
柔軟で全面で接触可能な変形筋外装により環境接触適応性を高めることができたと考えられる.

一方で軌道によっては\figref{fig:path}のように, 腱の経路が関節部から脱落してしまい, ある方向への動作が困難となる場合があった.
本研究では疑似球関節のモジュールを用いたが, 三次元的な構成で筋腱複合体駆動を用いる際は, 可動域を考慮した上で関節部分の構造も含めた設計が必要である.
}%

\section{Conclusion} \label{sec:conclusion}
\switchlanguage%
{%
This study developed a new muscle exterior to use the previously proposed wire-wound Muscle-Tendon Complex (ww-MTC) Drive in a three-dimensional body configuration. %\cite{MTC-2D:Ribayashi:HUMANOIDS2023}
The deformation is controlled by the internal frame of the PP sheets and covered by a braided sleeve to realize a deformable muscle exterior that can be in contact with the entire surface.
The consistency with theoretical calculations is verified, and it is shown that the number and size of the sheets can adjust the muscle's deformation.
In addition, a tendon as a thin and strong elastic element was constructed based on the previously proposed Stiffness Adjustable Tendon (SAT). %\cite{robomech2003:shirai:sat}

%We applied these muscle exteriors and tendon elements to a 2-axis 3-muscle robot capable of three-dimensional movements and verified their effectiveness.
We applied these muscle exteriors and tendon elements to a 2-axis 3-muscle robot and verified their effectiveness in a three-dimensional workspace.
The muscle exterior protected the robot's inner part and allowed it to adapt to environmental contact while continuing its movements flexibly.
It is concluded that the muscle exterior and tendon elements in this study can be used in a three-dimensional configuration and enhance the adaptability of wire-driven robots to environmental contact.

In the future, we will realize various muscle shapes existing in the human body and establish a body configuration method using muscle-tendon complexes together with the design of joint parts.
Furthermore, we aim to develop a humanoid robot suitable for environmental contact with its whole body and realize various motions by using the muscle-tendon complex drive.
}%
{%
  本研究では既に提案されていたワイヤ巻取式筋腱複合体駆動\cite{MTC-2D:Ribayashi:HUMANOIDS2023}を三次元的な身体構成においても使用するため, 新たな筋外装を開発した.
%膨張を伴う変形はアーチ状構造のたわみを利用している.
%PPシートの個数と大きさを変化させることで長さと幅の関係を調整した内部骨格を, 編組スリーブによって覆うことで, 全体で接触できる変形筋外装を実現した.
PPシートの内部骨格により変形を規定し, 編組スリーブによって覆うことで全体で接触できる変形筋外装を実現した.
筋外装について理論的な計算との整合性を検証し, シートの個数とサイズにより筋の変形を調整できることを示した.
また, 細くて強い弾性要素としての腱を既に提案されていたSAT\cite{robomech2003:shirai:sat}を元にして構成した.

これらの筋外装, 腱要素を, 三次元的な動作が可能な2軸3筋ロボットに適用して効果を検証した.
筋外装はロボット内部を保護し, 環境接触に対して柔軟に馴染みながらも動作を継続可能であった.
%
%これより本研究で用いた筋外装と腱要素は筋腱複合体コンセプトの利点を有し, 三次元的構成での使用に適したものとなっていることが確認された.
これより本研究の筋外装と腱要素は三次元的構成で使用でき, ワイヤ駆動ロボットの環境接触適応性を高めると結論づけられた.

%今後は多様な形状を力学的特性の設計も含めて実現すると共に, 関節部の設計と合わせて筋腱複合体による身体構成手法を確立していく.
今後は身体に存在する様々な筋肉の形状を実現し, 関節部の設計と合わせて筋腱複合体による身体構成手法を確立していく.
更に筋腱複合体駆動を用いて, 全身を有する環境接触に適したヒューマノイドの開発と多様な動作の実現を目指す.

}%

{
  %\footnotesize
  %\small
  %\bibliographystyle{junsrt}
  \balance
  \bibliographystyle{IEEEtran}
  \bibliography{bib}

\begin{thebibliography}{10}
\providecommand{\url}[1]{#1}
\csname url@rmstyle\endcsname
\providecommand{\newblock}{\relax}
\providecommand{\bibinfo}[2]{#2}
\providecommand\BIBentrySTDinterwordspacing{\spaceskip=0pt\relax}
\providecommand\BIBentryALTinterwordstretchfactor{4}
\providecommand\BIBentryALTinterwordspacing{\spaceskip=\fontdimen2\font plus
\BIBentryALTinterwordstretchfactor\fontdimen3\font minus
  \fontdimen4\font\relax}
\providecommand\BIBforeignlanguage[2]{{%
\expandafter\ifx\csname l@#1\endcsname\relax
\typeout{** WARNING: IEEEtran.bst: No hyphenation pattern has been}%
\typeout{** loaded for the language `#1'. Using the pattern for}%
\typeout{** the default language instead.}%
\else
\language=\csname l@#1\endcsname
\fi
#2}}

\bibitem{MTC-2D:Ribayashi:HUMANOIDS2023}
Y.~Ribayashi, K.~Miyama, A.~Miki, K.~Kawaharazuka, K.~Okada, K.~Kawasaki, and
  M.~Inaba, ``Development of a wire-wound muscle-tendon complex drive and its
  application to a two-dimensional robot configuration,'' in \emph{Proceedings
  of the 2023 IEEE-RAS International Conference on Humanoid Robots}, 2023, pp.
  758--764.

\bibitem{GR-1}
\BIBentryALTinterwordspacing
{Fourier Intelligence}, ``{GR-1},'' accessed: 2024-07-14. [Online]. Available:
  \url{https://fourierintelligence.com/gr1/}
\BIBentrySTDinterwordspacing

\bibitem{talos}
\BIBentryALTinterwordspacing
{PAL Robotics}, ``{TALOS},'' accessed: 2024-07-14. [Online]. Available:
  \url{https://pal-robotics.com/robots/talos/}
\BIBentrySTDinterwordspacing

\bibitem{apollo}
\BIBentryALTinterwordspacing
{APPTRONIK}, ``{APOLLO},'' accessed: 2024-07-14. [Online]. Available:
  \url{https://apptronik.com/apollo}
\BIBentrySTDinterwordspacing

\bibitem{neo}
\BIBentryALTinterwordspacing
{1X}, ``{NEO},'' accessed: 2024-07-14. [Online]. Available:
  \url{https://www.1x.tech/androids/neo}
\BIBentrySTDinterwordspacing

\bibitem{hiraoka2019whole}
N.~Hiraoka, M.~Murooka, H.~Ito, I.~Yanokura, K.~Okada, and M.~Inaba,
  ``Whole-body control of humanoid robot in 3d multi-contact under contact
  wrench constraints including joint load reduction with self-collision and
  internal wrench distribution,'' in \emph{2019 IEEE/RSJ International
  Conference on Intelligent Robots and Systems (IROS)}.\hskip 1em plus 0.5em
  minus 0.4em\relax IEEE, 2019, pp. 3860--3867.

\bibitem{ruscelli2020multi}
F.~Ruscelli, M.~P. Polverini, A.~Laurenzi, E.~M. Hoffman, and N.~G. Tsagarakis,
  ``A multi-contact motion planning and control strategy for physical
  interaction tasks using a humanoid robot,'' in \emph{2020 IEEE/RSJ
  International Conference on Intelligent Robots and Systems (IROS)}.\hskip 1em
  plus 0.5em minus 0.4em\relax IEEE, 2020, pp. 3869--3876.

\bibitem{ferrari2023multi}
P.~Ferrari, L.~Rossini, F.~Ruscelli, A.~Laurenzi, G.~Oriolo, N.~G. Tsagarakis,
  and E.~M. Hoffman, ``Multi-contact planning and control for humanoid robots:
  Design and validation of a complete framework,'' \emph{Robotics and
  Autonomous Systems}, vol. 166, p. 104448, 2023.

\bibitem{pratt2006capture}
J.~Pratt, J.~Carff, S.~Drakunov, and A.~Goswami, ``Capture point: A step toward
  humanoid push recovery,'' in \emph{2006 6th IEEE-RAS international conference
  on humanoid robots}.\hskip 1em plus 0.5em minus 0.4em\relax Ieee, 2006, pp.
  200--207.

\bibitem{stephens2010push}
B.~J. Stephens and C.~G. Atkeson, ``Push recovery by stepping for humanoid
  robots with force controlled joints,'' in \emph{2010 10th IEEE-RAS
  International conference on humanoid robots}.\hskip 1em plus 0.5em minus
  0.4em\relax IEEE, 2010, pp. 52--59.

\bibitem{semwal2017robust}
V.~B. Semwal, K.~Mondal, and G.~C. Nandi, ``Robust and accurate feature
  selection for humanoid push recovery and classification: deep learning
  approach,'' \emph{Neural Computing and Applications}, vol.~28, pp. 565--574,
  2017.

\bibitem{wittmeier2013toward}
S.~Wittmeier, C.~Alessandro, N.~Bascarevic, K.~Dalamagkidis, D.~Devereux,
  A.~Diamond, M.~J{\"a}ntsch, K.~Jovanovic, R.~Knight, H.~G. Marques,
  P.~Milosavljevic, B.~Mitra, B.~Svetozarevic, V.~Potkonjak, R.~Pfeifer,
  A.~Knoll, and O.~Holland, ``{Toward Anthropomimetic Robotics: Development,
  Simulation, and Control of a Musculoskeletal Torso},'' \emph{Artificial
  Life}, vol.~19, no.~1, pp. 171--193, 2013.

\bibitem{jantsch2013anthrob}
M.~J{\"a}ntsch, S.~Wittmeier, K.~Dalamagkidis, A.~Panos, F.~Volkart, and
  A.~Knoll, ``{Anthrob - A Printed Anthropomimetic Robot},'' in
  \emph{Proceedings of the 2013 IEEE-RAS International Conference on Humanoid
  Robots}, 2013, pp. 342--347.

\bibitem{iros2019:kawaharazuka:musashi}
K.~Kawaharazuka, S.~Makino, K.~Tsuzuki, M.~Onitsuka, Y.~Nagamatsu, K.~Shinjo,
  T.~Makabe, Y.~Asano, K.~Okada, K.~Kawasaki, and M.~Inaba, ``{Component
  Modularized Design of Musculoskeletal Humanoid Platform Musashi to
  Investigate Learning Control Systems},'' in \emph{Proceedings of the 2019
  IEEE/RSJ International Conference on Intelligent Robots and Systems}, 2019,
  pp. 7294--7301.

\bibitem{Humanoids2012:mizuuchi:pneumatic}
I.~Mizuuchi, M.~Kawamura, T.~Asaoka, and S.~Kumakura, ``Design and development
  of a compressor-embedded pneumatic-driven musculoskeletal humanoid,'' in
  \emph{2012 12th IEEE-RAS International Conference on Humanoid Robots
  (Humanoids 2012)}, 2012, pp. 811--816.

\bibitem{AR2018:hitzmann:anthropomorphic}
A.~Hitzmann, H.~Masuda, S.~Ikemoto, and K.~Hosoda, ``Anthropomorphic
  musculoskeletal 10 degrees-of-freedom robot arm driven by pneumatic
  artificial muscles,'' \emph{Advanced Robotics}, vol.~32, no.~15, pp.
  865--878, 2018.

\bibitem{TRO1996:Ching:McKibben}
C.-P. Chou and B.~Hannaford, ``Measurement and modeling of mckibben pneumatic
  artificial muscles,'' \emph{IEEE Transactions on Robotics and Automation},
  vol.~12, no.~1, pp. 90--102, 1996.

\bibitem{Act2022:Kalita:McKibben}
B.~Kalita, A.~Leonessa, and S.~K. Dwivedy, ``A review on the development of
  pneumatic artificial muscle actuators: Force model and application,''
  \emph{Actuators}, vol.~11, no.~10, p. 288, 2022.

\bibitem{iros2015:asano:module}
Y.~Asano, T.~Kozuki, S.~Ookubo, K.~Kawasaki, T.~Shirai, K.~Kimura, K.~Okada,
  and M.~Inaba, ``A sensor-driver integrated muscle module with high-tension
  measurability and flexibility for tendon-driven robots,'' in \emph{2015
  IEEE/RSJ International Conference on Intelligent Robots and Systems (IROS)},
  2015, pp. 5960--5965.

\bibitem{robomech2003:shirai:sat}
T.~Shirai and T.~Tomioka, ``Proposal of joint stiffness adjustment mechanism
  {SAT} -analysis and modeling of {SAT}-,'' in \emph{Proceedings of JSME
  Conference on Robotics and Mechatronics (ROBOMEC '03)}, 2003, pp.
  2P2--2F--F1, (in japanese).

\bibitem{1962:frisch:bar}
R.~Frisch-Fay, \emph{Flexible Bars}.\hskip 1em plus 0.5em minus 0.4em\relax
  Butterworths, London., 1962.

\bibitem{AR1997:koganezawa:antagonistic}
K.~Koganezawa, Y.~Watanabe, and N.~Shimizu, ``Antagonistic muscle-like actuator
  and its application to multi-dof forearm prosthesis,'' \emph{Advanced
  Robotics}, vol.~12, no. 7-8, pp. 771--789, 1997.

\bibitem{book2009:tank:anatomy}
P.~W. Tank and T.~R. Gest, \emph{Lippincott Williams \& Wilkins Atlas of
  Anatomy}.\hskip 1em plus 0.5em minus 0.4em\relax Butterworths, London., 2009.

\end{thebibliography}
}

\end{document}